\def\year{2022}\relax
\documentclass[letterpaper]{article} 
\pdfoutput=1 
\usepackage{aaai22}  
\usepackage{times}  
\usepackage{helvet}  
\usepackage{courier}  
\usepackage[hyphens]{url}  
\usepackage{graphicx} 
\usepackage{natbib}  
\usepackage{caption} 
\DeclareCaptionStyle{ruled}{labelfont=normalfont,labelsep=colon,strut=off} 
\usepackage{algorithm}
\usepackage{algorithmic}

\usepackage{hyperref}

\usepackage{newfloat}
\usepackage{listings}
\floatstyle{ruled}
\newfloat{listing}{tb}{lst}{}
\floatname{listing}{Listing}
\usepackage[utf8]{inputenc} 
\usepackage{booktabs}       
\usepackage{amsfonts}       
\usepackage{nicefrac}       
\usepackage{microtype}      
\usepackage{xcolor}         

\usepackage{comment}
\usepackage{todonotes}
\usepackage{bbm}
\usepackage{mathtools}
\usepackage{verbatim}
\usepackage[linesnumbered, ruled,vlined, algo2e]{algorithm2e}
\usepackage{enumitem}

\usepackage{nccmath}
\usepackage{soul} 
\usepackage{multirow}
\usepackage{gensymb} 

\usepackage{tikz}
\usetikzlibrary{intersections,shapes.arrows}

\definecolor{richardcolor}{rgb}{0.2, 0.4, 0.0}

\usepackage{amsthm,amsmath,amssymb, amsfonts}
\newtheorem{theorem}{Theorem}[section]

\newtheorem{observation}[theorem]{Observation}

\newcommand{\R}{\mathbb{R}}
\newcommand{\N}{\mathbb{N}}

\allowdisplaybreaks

\usepackage{amsmath}

\newcommand{\scom}[1]{\textcolor{Red}{\textit{[SK: #1]}}}

\title{Same State, Different Task: \\ Continual Reinforcement Learning without Interference}
\author {
    Samuel Kessler,\textsuperscript{\rm 1}
    Jack Parker-Holder, \textsuperscript{\rm 1}
    Philip Ball, \textsuperscript{\rm 1}
    Stefan Zohren, \textsuperscript{\rm 1}
    Stephen J. Roberts, \textsuperscript{\rm 1}
}
\affiliations {
    \textsuperscript{\rm 1} University of Oxford\\
    \texttt{\{skessler, jackph, ball, zohren, sjrob\}@robots.ox.ac.uk}
}

\begin{document}

\maketitle

\begin{abstract}
 Continual Learning (CL) considers the problem of training an agent sequentially on a set of tasks while seeking to retain performance on all previous tasks. A key challenge in CL is catastrophic forgetting, which arises when performance on a previously mastered task is reduced when learning a new task. While a variety of methods exist to combat forgetting, in some cases tasks are fundamentally incompatible with each other and thus cannot be learnt by a single policy. This can occur, in reinforcement learning (RL) when an agent may be rewarded for achieving \emph{different goals} from the \emph{same observation}. In this paper we formalize this ``interference'' as distinct from the problem of forgetting. We show that existing CL methods based on single neural network predictors with shared replay buffers fail in the presence of interference. Instead, we propose a simple method, OWL, to address this challenge. OWL learns a factorized policy, using \emph{shared} feature extraction layers, but \emph{separate} heads, each specializing on a new task. The separate heads in OWL are used to prevent interference. At test time, we formulate policy selection as a multi-armed bandit problem, and show it is possible to select the best policy for an \emph{unknown task} using feedback from the environment. The use of bandit algorithms allows the OWL agent to constructively re-use different continually learnt policies at different times during an episode. We show in multiple RL environments that existing replay based CL methods fail, while OWL is able to achieve close to optimal performance when training sequentially.
\end{abstract}

\section{Introduction}



Reinforcement Learning (RL \cite{sutton1998introduction}) considers the problem of an agent taking sequential actions in an environment to maximize some notion of reward. In recent times there has been tremendous success in RL, with impressive results in Games \cite{alphago} and Robotics \cite{dexterity}, and even real-world settings \cite{loon}. However, these successes have predominantly focused on learning a \textit{single} task, with agents often brittle to changes in the dynamics or rewards (or even the seed \cite{deeprlmatters}). 

\begin{figure}[h]
    \centering
    \includegraphics[width=1.0\linewidth]{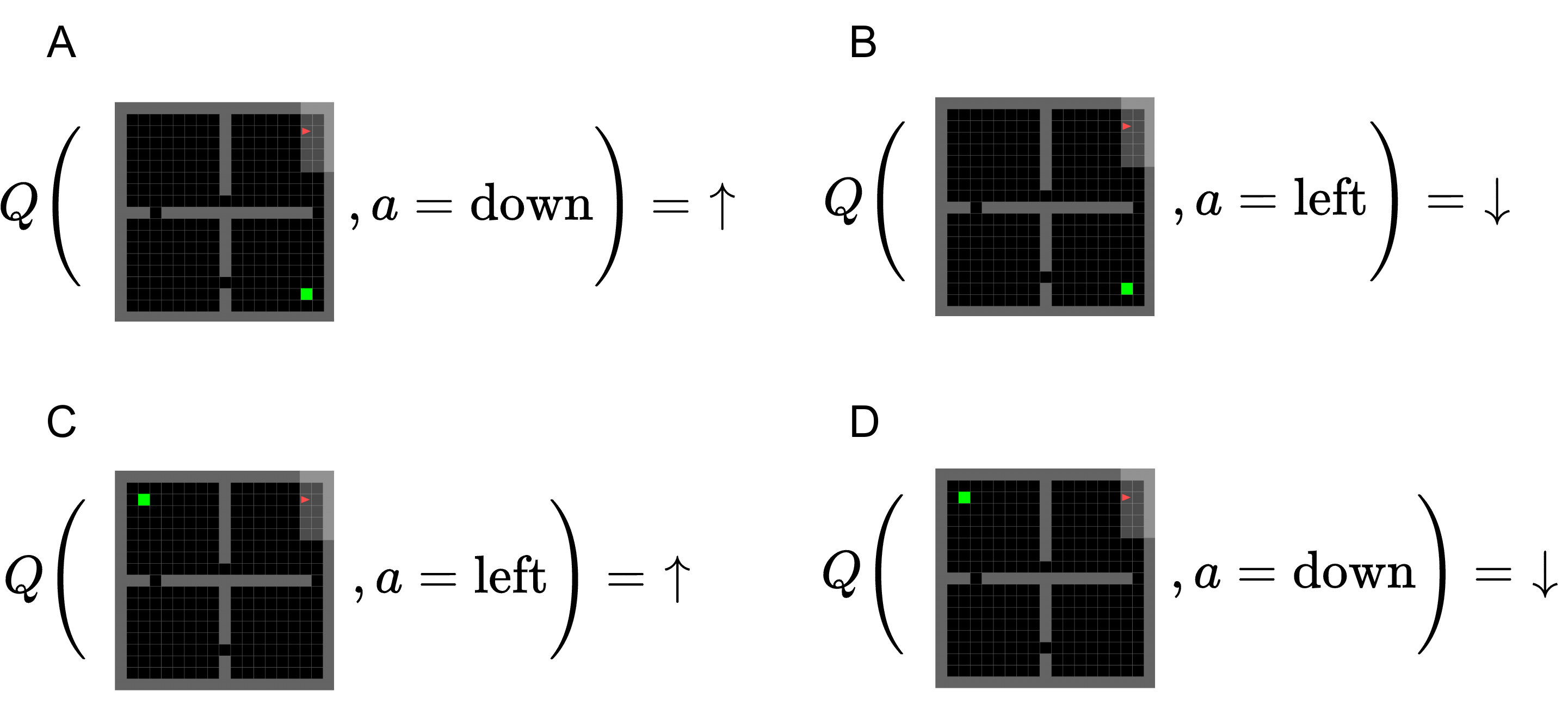} 
    \caption{\small{\textbf{A} and \textbf{B} a simple four rooms environment task where the goal (the green square) is located at the bottom right. \textbf{C} and \textbf{D} a different four rooms task where the goal is at a different location, the optimal Q-function should give different values for the same action for the same starting state. The symbol $\uparrow$ denotes a high Q-value and $\downarrow$ a low Q-value. Task agnostic, single predictor continual RL methods will be sub-optimal compared to methods which use the task identifier to condition their policies like OWL.}}
    \label{fig:interference_4R}
    \vspace{-2mm}
\end{figure}

One of the most appealing qualities of RL agents is their ability to continue to learn and thus improve throughout their lifetime. As such, there has recently been an increase in interest in \emph{Continual Reinforcement Learning} (CRL), \cite{ring1994continual, crl_review}, a paradigm where agents train on tasks sequentially, while seeking to maintain performance on previously mastered tasks \cite{ewc, progress_compress, clear}. A key issue when training on sequential tasks is \emph{catastrophic forgetting}, a consequence of gradient based learning, as training on a new task overwrites parameters which were important for previous tasks \cite{RobertM.French1999}. While existing methods address this, they typically only consider the setup where each ``task'' is an entirely different RL environment, for example different games from the Arcade Learning Environment \cite{ale} or different simulated robotics environments \cite{Ahn2019}.

In this paper, we focus on a more challenging problem. Instead of distinct environments as different tasks, we consider learning to solve different tasks with the \emph{same state space}. Since these tasks may not only have different but even opposite optimal actions for the same observation, training on them sequentially causes what we call ``interference'' which can in turn \emph{induce} forgetting, as the agent directly optimizes for an opposing policy. Interference can also lead to reduced performance for future tasks, as the agent cannot fully learn to solve a new task given it's retained knowledge of previous opposing objectives. This problem regularly arises in real-world settings, whereby there are a multitude of tasks with reward functions which all share the same state space (e.g. a visual observation of the world). Since previous CRL methods used different environments as different tasks then the agents can learn that the different state spaces correspond to different optimal behaviors and so interference is rarely exhibited.

We begin by showing a simple supervised setting where it is impossible to solve interfering tasks, continually with a single predictor despite using strong CL techniques to prevent forgetting, Figure~\ref{fig:multimnist}. This setting can occur in RL where we have different goals but the same observation for different tasks, see Figure~\ref{fig:interference_4R}. We therefore introduce a new approach to CRL, which we call C\textbf{\underline{O}}ntinual RL \textbf{\underline{W}}ithout Conf\textbf{\underline{L}}ict or OWL. OWL makes use of \emph{shared feature extraction layers}, while acting based on \emph{separate independent policy heads}. The use of the shared layers means that the size of the model scales gracefully with the number of tasks. The separate heads act to model multi-modal objectives and not as a means to retain task specific knowledge, as they are implicitly used in CL. Task agnostic, single predictor CRL methods which use experience replay \cite{clear} will thus suffer from this interference. But have the advantage of not having to infer the task the agent is in to then solve it. Experience replay has been shown to be a strong CL strategy \cite{balaji2020effectiveness}. 

In the presence of interference, task inference is now necessary and at test time, OWL adaptively selects the policy using a method inspired by multi-armed bandits. We demonstrate in a series of simple yet challenging RL problems that OWL can successfully deal with interference, where existing methods fail. To alleviate forgetting we consider simple weight space \cite{ewc} and functional regularizations \cite{hinton2015distilling} as these have been repeatedly shown to be effective in CRL \cite{lifelonghanabi}. 

Our core contribution is to identify a challenging new setting for CRL and propose a simple approach to solving it. As far as we are aware, we are the first to consider a bandit approach for selecting policies for deployment, inferring unknown tasks. The power of this method is further demonstrated in a set of generalization experiments, where OWL is able to solve tasks it has never seen before, up to $6$x more successfully than the experience replay baseline \cite{clear}.

\section{Related Work}
\label{sec:related}

\textbf{Continual Learning in Supervised Learning:} Continual Learning (CL) is a sequential learning problem. One approach to CL, referred to as \emph{regularization approaches} regularizes a NN predictor's weights to ensure that new learning produces weights which are similar to previous tasks \cite{ewc, vcl, Zenke2017}. Alternatively, previous task functions can be regularized to ensure that the functions mapping inputs to outputs are remembered \cite{Li2017}. By contrast, \emph{expansion approaches} add new neural resources to enable learning new tasks while preserving components for specific tasks \cite{progressivenets, SoochanLeeJunsooHaDongsuZhang2020}. \emph{Memory approaches} replay data from previous tasks when learning the current task. This can be performed with a generative model \cite{Shin2017}. Or small samples from previous tasks (\emph{memories}) \cite{Lopez-Paz, Aljundi2019, ChaudhryTinyEps}. Meta-learning pre-training has been explored to augment continual learning (and \emph{context-dependent targets} can allow interference but is not explored) \cite{caccia2020online}.

\textbf{Continual Learning in Reinforcement Learning:} The CL regularization method EWC has been applied to DQN \cite{Mnih} to learn over a series of Atari games  \cite{ewc}. 
Both Progressive Networks \cite{progressivenets} and Progress and Compress \cite{progress_compress} are applied to policy and value function feature extractors for an  actor-critic approach. These methods are told when the task changes.

CLEAR leverages experience replay buffers \cite{experience_replay} only to prevent forgetting: by using an actor-critic with V-trace importance sampling \cite{espeholt2018impala} of past experiences from the replay buffer catastrophic forgetting can be overcome \cite{clear}. CLEAR uses a single predictor NNs and all experience is stored in the replay buffer and thus CLEAR is not required to know when the task changes. However, interference is ignored as multi-task performance of the all environments is similar to the sum of performances of individual environments. Different selective experience replay strategies can be used for preserving performance on past tasks \cite{Isele2018}. Alternatively, \cite{Mendez2020} learns a policy gradient model which factorizes into task specific parameters and shared parameters. OWL is more general as it can wrap around any RL algorithm and be used for discrete action spaces and continuous control settings and achieve better results.

A number of previous works have studied transfer in multi-task RL settings where the goals within an environment change \cite{barreto2016successor, schaul2015universal}. Our work is related to \cite{Yu2020} which considers interference between gradients in a multi-task setting.


\textbf{Interference:} this problem is discussed in \cite{clear} and has been studied in multi-task (for example \cite{bishop2012bayesian, lin2019pareto}) and meta-learning (for example, \cite{meta_augmentation}). However, we believe we are the first to consider it in CRL, and that the current state-of-the-art methods lack an approach to tackle it. In particular, we note that existing replay based methods such as \cite{clear} fail to address this issue, as the experience replay buffer will contain tuples of the same state-action pairs but different rewards for different tasks. Thus, the agent will not converge, as we show later in our experiments.

\section{Background}
\label{sec:background}

\subsection{Reinforcement Learning}

A Markov Decision Process ($\mathrm{MDP}$, \cite{bellmanmdp}) is a tuple $(\mathcal{S},\mathcal{A},P,R, \gamma)$. Here $\mathcal{S}$ and $\mathcal{A}$ are the sets of states and actions respectively, such that for $s_t, s_{t+1} \in \mathcal{S}$ and $a_t \in \mathcal{A}$. $P(s_{t+1}| s_t, a_t)$ is the probability that the system/agent transitions from $s_t$ to $s_{t+1}$ given action $a_t$ and $R(a_t,s_t,s_{t+1})$ is a reward obtained by an agent transitioning from $s_t$ to $s_{t+1}$ via $a_t$. The discount factor is represented by $\gamma \in (0,1)$. Actions $a_t$ are chosen using a policy $\pi$ that maps states to actions: $\mathcal{S}\rightarrow\mathcal{A}$. In this paper we consider MDPs with finite horizons $H$. The return from a state is defined as the sum of discounted future rewards $R_t = \sum_{i=t}^H \gamma^{(i \text{-} t)}r(s_i, a_i)$. In RL the objective is to maximize $J = \mathbb{E}_{a_i \sim \pi}[R_1|s_0]$ given an initial state $s_0$, sampled from the environment.

One approach to maximizing expected return is to learn an action-value function for each state-action pair: $Q^\pi(s_t,a_t) = \mathbb{E}_{s_t \sim P, r_t\sim R, a_t \sim \pi}[\sum_t R_t]$. We parameterize the action-value function with a neural network, denoted $Q(s_t, a_t; \theta_i)$, with parameters $\theta_i$. We optimize $\theta_i$ to minimize the expected temporal difference error using gradient descent:
\begin{align}
\label{eq:td-error}
    \mathcal{L}_i(\theta_i) = \mathbb{E}_{s_t, a_t \sim \rho}[(y_i - Q(s_t, a_t; \theta_i))^2]
\end{align}
We use Q-learning \cite{Watkins1992}, giving $y_i = r_t + \gamma \mathrm{max}_a Q(s_{t+1}, a; \theta_{i-1})$, and the parameters $\theta_{i-1}$ are target network parameters. The associated update does not require on-policy samples, so learning is off-policy using a replay buffer \cite{experience_replay}, hence $\rho$ is an empirical distribution that represents samples from the buffer. For continuous action settings, we learn a policy $\pi_\phi:\mathcal{S} \rightarrow \mathcal{A}$ using a neural network parameterized by $\phi$. For discrete action settings, our policy follows the maximum Q-value: $\pi \coloneqq \text{argmax}_a Q(s,a)$.

\subsection{Continual Learning}
\emph{Continual learning} (CL) is a paradigm whereby an agent must learn a set of tasks sequentially, while maintaining performance across all tasks. This presents several challenges, in particular avoiding forgetting and efficiently allocating resources for learning new tasks. In CL, the model is shown $M$ tasks $\mathcal{T}_{\tau}$ sequentially, where $\tau= 1, \ldots M$. A task is defined as $\mathcal{T}_\tau = \big\{p(X), p(Y| X), \tau\big\}$ where $X$ and $Y$ are input and output random variables.

{ \centering
\begin{figure}[h]
    \centering
    \includegraphics[width=1.0\linewidth]{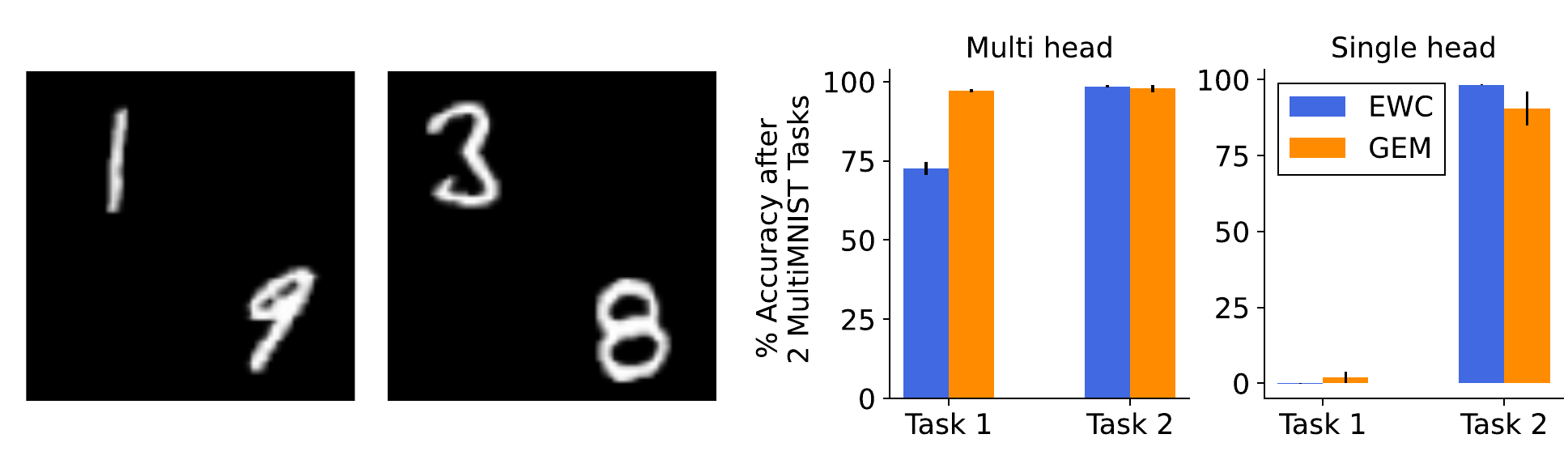} 
    \caption{\small{\textbf{Left} two samples from the MultiMNIST dataset \cite{sabour2017dynamic}. The first task $\mathcal{T}_1$ requires classifying the top left digit and the second task $\mathcal{T}_2$ requires classifying the bottom right. \textbf{Right} results from continually learning on both MultiMNIST tasks. We compare two different CL methods EWC \cite{ewc} and GEM \cite{Lopez-Paz}. Single head task agnostic setups suffer from catastrophic forgetting of the first task due to interference. The multi-head setup is a simple remedy to the problem of interference.}}
    \label{fig:multimnist}
\end{figure}
}

In practice a task is comprised of $i=1 \ldots N_{\tau}$ inputs $x_i \in \R^d$ and outputs $y_i \in \R$ ($\N \subset \R$). The model will lose access to the training dataset for task $\mathcal{T}_\tau$, it will be continually evaluated on all previous tasks $\mathcal{T}_j$ for $j \leq \tau$. $\tau$ can be used as a task identifier informing the agent when to start training on a new task. For a comprehensive review of CL scenarios see \cite{VanDeVen}. For RL the definition of a task is simply an MDP and task identifier: $\mathcal{T}_{\tau} = \big\{\mathcal{S}_{\tau}, \mathcal{A}_{\tau}, p_{\tau}(s_1), p_{\tau}(s_{t+1}|s_{t}, a_t), R_{\tau}(a_t, s_t, s_{t+1}), {\tau} \big\}$, and so the agent will no longer be able to interact with previous environments, but must ensure that it can remember how to solve all past tasks/environments.

\section{Catastrophic Forgetting vs. Interference}
\label{sec:motiv}

Forgetting occurs when performance on old tasks is reduced while learning new tasks. On the other hand, interference occurs when two or more tasks are incompatible for the same model. We re-use these definitions from CLEAR \cite{clear}. We observe this when the multi-task objectives are multi-modal and tasks share the same observation space but have different goals/objectives.

We demonstrate interference using the MultiMNIST dataset \cite{sabour2017dynamic}, Figure~\ref{fig:multimnist}. Each image is composed of two different MNIST digits and for $\mathcal{T}_1$ we are required to classify the top image and in $\mathcal{T}_2$ we are required to classify the bottom image. For both of these tasks the only difference is the objective. When we perform CL with a single predictor network or single-headed network with different CL strategies to alleviate forgetting we see that the interference between tasks causes almost $100\%$ forgetting of the first task, despite using established CL strategies. On the other hand using multi-headed networks allows us to model both objectives in MultiMNIST. We observe that we will get interference in the following.

\begin{observation}
\label{def:sl_interference_function_level}
Consider two tasks $\mathcal{T}_i$ and $\mathcal{T}_j$. Let both tasks' input distributions $p_k(X)$ share the same support but have different conditional distributions $p_k(Y|X)= \mathcal{N}(f^{k}(X), \beta^{-1})$, where $f^k$ is a mean function with $f^i \ne f^j$ and $\beta^{-1}$ is data noise. Then the multi-task distribution is bi-modal and using a Gaussian likelihood will result in interference.
\end{observation}

This may seem contrived in the supervised setting, however, it is common throughout reinforcement learning. Consider a partially observable MDP (POMDP) where we receive an initial observation but do not know the goal location or reward function then an agent might require different policies for each task. We see an example of this in Figure~\ref{fig:interference_4R} where in one task the goal is in the room below the agent and in the other the goal is in the room to the left. The most efficient policies will guide the agent in different directions depending on the task the agent is in. This observation has important consequences: methods which are task agnostic and do not condition on the task or do not use task specific parameters are susceptible to interference. Some CL methods use a single predictor and aim to approximate the multi-task setting by using storing samples in a buffer \cite{aljundi2019online, clear}. Furthermore single-headed networks are often used as more difficult CL scenarios when studying methods to mitigate forgetting \cite{VanDeVen, farquhar2018towards}. 

Our solution is to model the multi-modality by learning a mixture of linear regressions (see: §14.5.1 \cite{bishop2006pattern}). The same applies to CRL: the Q-function needs to have separate weights for each task or needs to condition on the task to solve CRL environments with interference. In practice we will use multi-headed network predictors. \emph{Whereas these are commonly employed to preserve previous task knowledge and prevent forgetting in CL, we are employing them as a means to prevent interference}. So our motivation for the use of multi-head networks is wholly different. Most supervised CL settings are benchmarked on vision tasks which use different distinct classes as tasks. Thus $p_{\tau}(X)$ and $p_{\tau}(Y|X)$ both change for each $\tau$; a single NN predictor can model this. However in RL only the reward function, $R_{\tau}(s_{t}, a_{t}, s_{t+1})$, need change from one task to another. 

\section{Continual RL without Conflict}
\label{sec:alg}


\begin{figure}
    \centering
    \vspace{-0mm}
    \includegraphics[width=.95\linewidth]{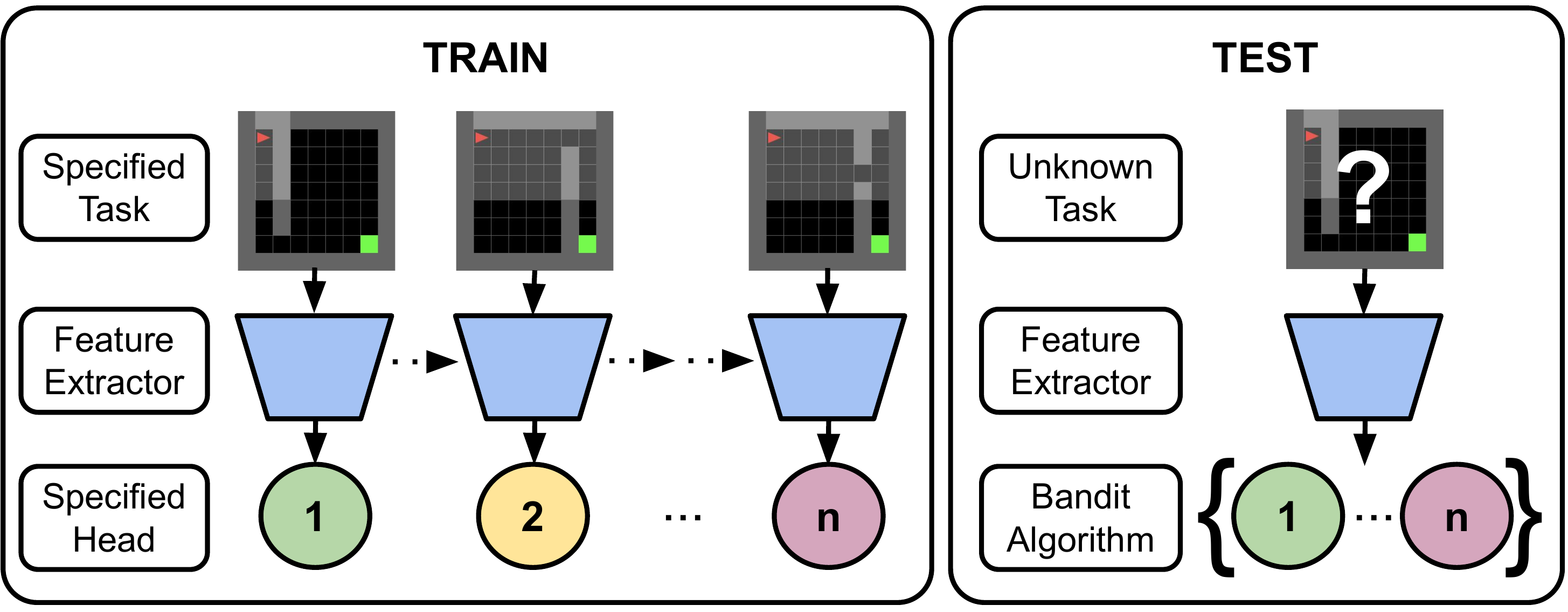}
    \caption{\small{An overview of our approach. On the left we show the training setup. When training sequentially on different tasks we use the same feature extractor (blue), regularized with EWC, but maintain a set of distinct heads (green, yellow, ..., pink) corresponding to different tasks. On the right we show the test time adaptation. Note that the task is not known, and must be inferred through interaction with the environment.}}
    \label{figure:unclear}
    \vspace{-6mm}
\end{figure}

At a high level, OWL uses an off-policy RL algorithm to train a Q-function across tasks sequentially. To prevent interference our key insight is that:  1.) we can use a single network with a shared feature extractor but multiple heads, parameterized by linear layers to fit individual tasks; 2.) we flush the experience replay buffer when starting to learn in a new task. At test time, we frame policy/head selection as a multi-armed bandit problem, to adaptively select the best policy. In this section we provide additional details on each component, describing first the structure of the Q-function, before moving to our test time adaptation.  

\subsection{Factorized Q-Functions}

Multi-head networks are commonly used in CL \cite{Li2017, vcl}, they enable learning task specific mapping from a shared feature extractor to the output layers. Multi-head networks are effective in that allow learning of task specific parameters which can we recalled and so help to alleviate forgetting. Single-head networks are commonly used as a more difficult baseline for CL benchmarks \cite{farquhar2018towards, VanDeVen}. In our work multi-head networks are used as they prevent interference.

{\centering
\begin{minipage}{.99\linewidth}
    \begin{algorithm}[H]
    \textbf{Input:} Tasks $ \mathcal{T} = \{\mathcal{T}_i\}_{i=1}^M$.\\
    \textbf{Initialize:} $\theta$ and $\phi$, $\Omega^{Q} = \Omega^{\pi} = \emptyset$. \; \\
    \For{$t = 1, 2, \ldots, M$}{
        1. See Task $\mathcal{T}_t$\; \\
        2. Train Q-function with parameters $\{\theta_z, \theta_i\}$ and regularization $\Omega^{Q}$. \; \\
        \If{$\mathcal{A}$ is continuous}{
        3. Train policy with parameters $\{\phi_z, \phi_i\}$ with regularization $\Omega^{\pi}$. \;
        }
        4. Calculate Q-function EWC regularization and $\Omega^{Q} := \{\mathcal{L}^Q_{\textrm{EWC}}, \Omega^{Q} \}$. \; \\
        \If{$\mathcal{A}$ is continuous}{
        5. Calculate policy EWC regularization and $\Omega^{\pi} := \{\mathcal{L}^{\pi}_{\textrm{EWC}}, \Omega^{\pi} \}$. \;
        }
        6. Empty the experience reply buffer $\mathcal{D} = \emptyset$. \; \\
        7. Evaluate according to Algorithm~\ref{alg:unclear_test}. \;
     }
     \caption{OWL: Training}
    \label{Alg:unclear_train}
    \end{algorithm}
\vspace{3.0mm}
\end{minipage}
}

\textbf{Alleviating forgetting:} We represent a factorized Q-function as having parameters $\theta = \{\theta_z, \theta_{1:M}\}$, where $\theta_z$ are feature extractor parameters and $\theta_{1:M}$ the heads. For discrete problems one can follow the maximum Q-values to obtain the next action. For parameterized policies we can equally construct our policy similarly with parameters $\phi = \{\phi_z, \phi_{1:M}\}$ where $\phi_z$ are the neural network feature extraction layers, and $\phi_{1:M}$ are linear policy heads. To address forgetting in the shared neural network feature extractors we use regularization methods. In particular we found EWC to work well and is a simple approach to prevent forgetting \cite{ewc}, (we also tried a functional regularization \cite{hinton2015distilling, Li2017} but found it underperformed; see Section~\ref{app:ablations}). We train our agent according to Algorithm~\ref{Alg:unclear_train}. As we see more and more tasks new heads can easily be added and so we do not need to prespecify the number of tasks or policy heads $M \in \{1, \ldots, \infty \}$.

\subsection{Selecting Policies as a Multi-Armed Bandit Problem}

\begin{figure*}
    \centering
    \includegraphics[width=1.0\linewidth]{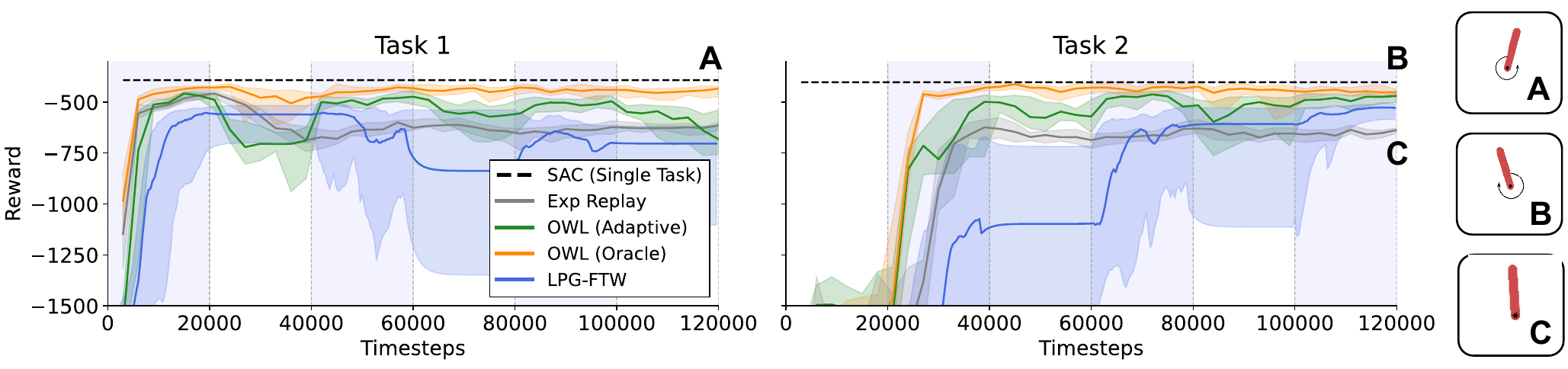} 
    \caption{\small{Median performance (with inter-quartile range) across $10$ seeds. Pale blue shaded regions correspond to the timesteps when the task in question is being trained on. \textbf{A}, Task 1 arm position from the OWL agent (oracle) \textbf{B}, Task 2 arm position from the OWL agent (oracle) \textbf{C}, Optimal arm position for Exp Replay for Tasks 1 and 2 (note that the reward function also has angular velocity terms thus the OWL/SAC agent isn't able to place the arms exactly at $\pm 90^{\circ}$ without obtaining a sub optimal reward). Exp Replay clearly displays interference.}}
    \vspace{-4.0mm}
    \label{fig:pendulum_main}
\end{figure*}

At test time we do not tell OWL which task it is being evaluated on. We consider the set of arms $M$ to be the set of policies which can be chosen to act at each timestep of the test task. The aim is to find the policy which achieves the highest reward on a given test task. We use a modified version of the Exponentially Weighted Average Forecaster algorithm \cite{cesa2006prediction}, as has been shown to be successful adapting components of RL algorithms \cite{rp1}. In this setup we consider $M$ experts making recommendations at the beginning of each round. After sampling a decision $i_t \in \{ 1, \cdots, M\}$ from a distribution $\mathbf{p}^t \in \Delta_M$ with the form $\mathbf{p}^t(i) \propto \exp\left(  \ell_t(i)   \right)$ the learner experiences a loss $l_{i_t}^t \in \mathbb{R}$. The distribution $\mathbf{p}^t$ is updated by changing $\ell_t$ as follows:
\vspace{-1mm}
\begin{equation}
\label{equation::exponential_weights_update}
    \ell_{t+1}(i) = \begin{cases}
        \ell_t(i) + \eta \frac{l_i^t}{\mathbf{p}^t(i)} & \text{if } i = i_t \\
        \ell_t(i) &\text{otherwise,}
        \end{cases}
\end{equation}
for some step size parameter $\eta$. We consider the case where the selection of $\phi_i$ is thought of as choosing among $M$ experts which we identify as the different policies $\{\phi_i \}_{i=1}^M$, trained on the corresponding Q-functions $\{\theta_i \}_{i=1}^M$. The loss we consider is of the form $l_{i_t} = 1 / \hat{G}_{\phi_{i}}(\theta_{i})$, where ${G}_{\phi_{i}}(\theta_{i})$ is the TD error or the log likelihood of the observed reward from the test task, $r_t$ given the predicted Q-values. If required, we can then perform a normalization of $G$, hence $\hat{G}$. Henceforth we denote by $\mathbf{p}^t_\phi$ the exponential weights distribution over $\phi$ values at time $t$. The pseudocode for our test-time procedure is shown in Algorithm \ref{alg:unclear_test}.

{\centering
\begin{minipage}{.99\linewidth}
    \centering\begin{algorithm}[H]
    \textbf{Input:} tasks seen so far $\mathcal{T} = \{\mathcal{T}_1, \ldots, \mathcal{T}_{\tau}\}$, Q-functions $\{\phi_i\}_{i=1}^{M}$, step size $\eta$, maximum number of timesteps $T$. \\
    \textbf{Initialize:} $\mathbf{p}_\phi^1$ as a uniform distribution, $s_1$ as the initial state of the test task. \; \\
    \For{$\mathcal{T}_j \in \mathcal{T}$}{
    \For{$t = 1, \ldots, T-1$}{
      1. Select $i_t \sim \mathbf{p}_\phi^t$, and set $\pi_\mathrm{test} = \pi_{\phi_{i_t}}$. \;\\
      2. Take action $a_t \sim \pi_\mathrm{test}(s_t)$, and receive reward $r_t$ and the next state $s_{t+1}$ from $\mathcal{T}_j$.\;\\
      3. Use  Equation \ref{equation::exponential_weights_update} to update  $\mathbf{p}_\phi^t$ with $l_{i_t}^t = \hat{G}_{\phi_t}(\theta_{t+1})$
     }
     }
     \caption{OWL: Testing}
    \label{alg:unclear_test}
    \end{algorithm}
\end{minipage}
}



\section{Experiments}
\label{sec:experiments}

To test our approach, we consider challenging CRL problems where tasks have similar or identical state and actions spaces but distinct goals/rewards. Our main hypothesis is that these tasks cannot be solved continually with a single policy, using a shared replay buffer. Our primary baseline, which we call Experience Replay (Exp Replay in figures) corresponds to this case \cite{clear} and has been shown to be a very effective baseline in CL \cite{balaji2020effectiveness}. In each setting we test two versions of our algorithm, which we refer to as $\mathrm{Oracle}$ and $\mathrm{Adaptive}$. With the $\mathrm{Oracle}$, the OWL agent is told at test time which task is being evaluated. Finally, we consider the multi-arm bandit (MAB) approach, denoted $\mathrm{Adaptive}$. Our code is available at \href{https://github.com/skezle/owl}{\textcolor{purple}{github.com/skezle/owl}}.

\subsection{Pendulums with Interfering Goals}

The first setting we consider is a simple yet challenging take on the well-known $\mathrm{Pendulum}$-$\mathrm{v0}$ environment \cite{gym}. Typically, the policy is rewarded for placing the pendulum at $0^{\circ}$. Instead, we amend the reward function to produce two interfering tasks, with optimal positions: $\{+90^{\circ}, -90^{\circ}\}$. We have continuous actions and train a multi-head policy and Q-function using Soft Actor Critic \cite{sac-v2}. We train on each task three times, switching every $20,000$ environment steps. 
For more details on our implementation see Section~\ref{sec:sac_implementation_details}. The results are shown in Fig. \ref{fig:pendulum_main}.


\begin{figure*}[h]
    \vspace{-0mm}
    \centering
    \includegraphics[width=1.0\linewidth]{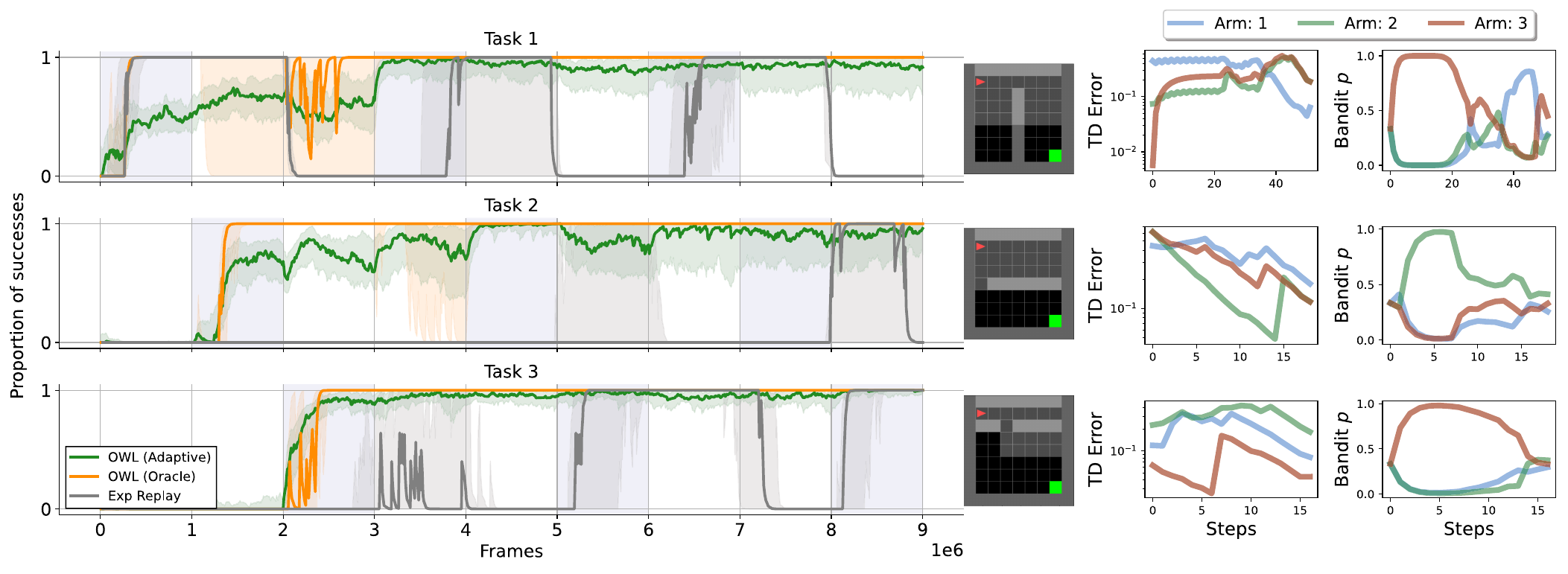} 
    \caption{\small{\textbf{Left.} Median performance across 10 seeds for three different MiniGrid environments trained continually. Shaded envelopes correspond to the inter-quartile range. Shaded pale blue regions corresponds to the current training task. OWL $\mathrm{Adaptive}$ and $\mathrm{Oracle}$ are able to prevent forgetting and interference while Exp. Replay fails. \textbf{Right.} Bandit arm probabilities over the course of a roll-out to demonstrate how the TD error feedback is used to select the right arm/policy to solve the task (note these have been smoothed for visualization purposes).}}
    \label{fig:minigrid_t3_ewc500}
    \vspace{-3mm}
\end{figure*}

First, we see evidence confirming our first hypothesis that training with a shared replay buffer over all tasks leads to suboptimal performance on both tasks due to the interfering nature of the tasks (Exp Replay). The Exp Replay agent (grey) learns to place the pendulum at $0^{\circ}$. As we see on the bottom right, this balances the conflicting goals, but is suboptimal for both individual tasks (see Fig.\ref{fig:pendulum_main}C). Secondly, the Oracle version of OWL (orange), which knows the task under evaluation at test time, performs well since two separate policies trained on the individual tasks (black, dashed) and displays minimal forgetting. Encouragingly we can almost achieve this same performance without informing the agent of the task index, using our adaptive mechanism (green). Our method also outperforms LPG-FTW \cite{Mendez2020} which uses $2.5$x more gradient steps as our method builds on top of SAC which yields state-of-the-art results in continuous control. Regarding feedback to the algorithm, we explore the importance of the probabilistic networks in the Appendix, Section~\ref{app:ablations}. 

\subsection{MiniGrid Environments}

MiniGrid is a challenging set of procedurally generated maze environments \cite{gym_minigrid}. Each environment is partially observable, with the agent only ``seeing'' a small region of visual input out of a larger state. Additionally, each state is an image, and rewards are sparse (the agent only receives a reward for navigating to the green tile in Figure~\ref{figure:unclear}), which makes it harder for agents to find learning signals. We use the $\mathrm{SimpleCrossing}$ environment, which has a single wall. Each environment seed corresponds to a different wall position, orientation and different door, thus generalization is challenging. The initial observation can look identical (or very similar) for two different environments, and the agent has to explore to discover the wall location and door position. 

We employ DQN to handle the discrete action space \cite{Mnih}, see Section~\ref{sec:dqn_implementation_details} for implementation details. We train the same methods as the previous experiment on three distinct MiniGrid grid worlds continually, repeating each three times for $1$M steps. We use the TD error in Eq~(\ref{eq:td-error}) as feedback to the MAB. OWL (Oracle) is able to consistently solve all environments after one round Figure~\ref{fig:minigrid_t3_ewc500}. OWL (Adaptive) is able to dynamically select the correct policy most of the time after seeing each task once, and continues to improve with training, with the final performance almost matching OWL (oracle)\footnote{For videos of OWL in action, see: \url{https://sites.google.com/view/crlwithoutconflict/}}. Exp Replay exhibits significant interference between tasks.

\begin{figure*}[h]
    \hspace*{-1cm}
    \centering
    \includegraphics[width=1.0\linewidth]{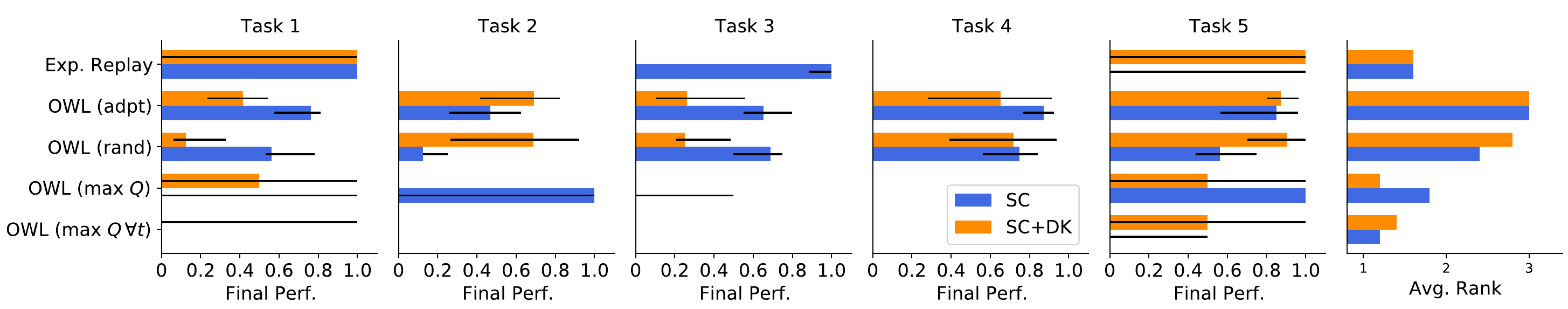} 
    \caption{\small{Final performance for different OWL policy selection strategies and Exp Replay.}}
    \label{fig:minigrid_t5_selection_strategies}
\end{figure*}


\textbf{Scaling to more tasks.} We now scale to $5$ $\mathrm{SimpleCrossing}$ tasks (denoted as SC in plots) and another set of $5$ tasks with $3$ $\mathrm{SimpleCrossing}$ and $2$ $\mathrm{DoorKey}$ environments (denoted as SC+DK). The set of tasks are repeated $3$ times each task is seen for $0.75$M environment steps. For Exp Replay we adjust the buffer size to $4$M ensure that data from all tasks are in the buffer over the course of training. We note that Exp. Replay again suffers from interference while OWL is able to overcome it (and forgetting) see Figures~\ref{fig:minigrid_t5_sc} and \ref{fig:minigrid_t5_sc_dk} in the appendix.

\begin{figure*}[h]
    \hspace*{-1cm}
    \centering
    \includegraphics[width=1.0\linewidth]{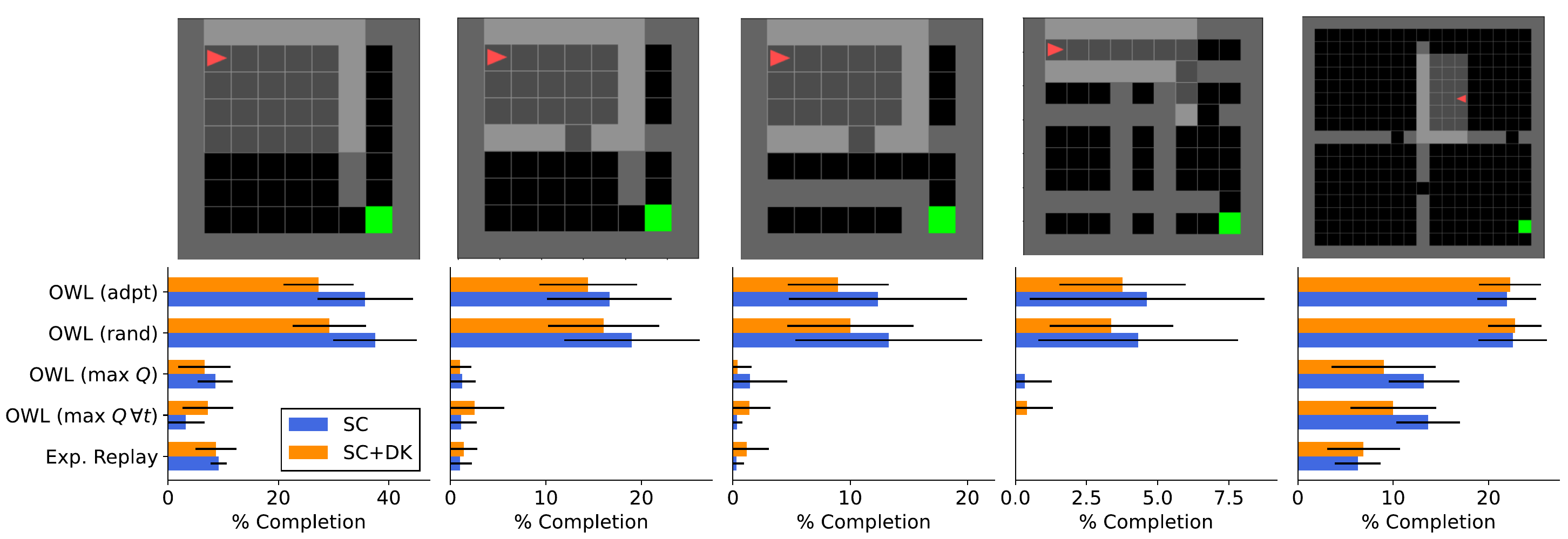} 
    \caption{\small{Mean and std error proportion of successes for $100$ different environments which have not been seen during training for $10$ seeds. OWL is able to generalize to unseen environments while Exp Replay fails.}}
    \label{fig:gen_arm_selection_strategies}
\end{figure*}

We explore different arm selection strategies in Figure~\ref{fig:minigrid_t5_selection_strategies}. We compare the multi-armed bandit (MAB) versus random policy head selection at each step of the roll-out (OWL (rand)), versus selecting the policy head with the largest expected reward (OWL (max Q)) and versus selecting the policy with the largest expected reward at each step of the roll-out (max Q$ \, \forall t$). We see that approaches which use the Q-value to select the policy fail, as does Exp Replay. Random policy selection performs well, but the MAB performs significantly better than random policy selection, a t-test with unequal variances has a $p$-value of $0.06$ for the SC policies, there isn't a significant difference for the SC+DK policies. Thus there are statistically significant benefits to using the MAB. The MAB approach can also behave similarly to a random policy head selection can perform well \cite{randomsearch, mania2018simple}.

\textbf{Generalization Results.} Many works have focused on generalization properties of RL agents in the MiniGrid environment \cite{primitives}, training on hundreds of levels. Instead, we train on just $5$ levels sequentially, which produces a significant risk of overfitting. We take the final SC and SC+DK policies and evaluate them on $100$ different, unseen tasks Figure~\ref{fig:gen_arm_selection_strategies}. We find that our OWL agent is able to transfer effectively to these unseen tasks, solving up to $40\%$ of unseen levels single walled levels, around $4\times$ more than Exp Replay. OWL can even solve harder environments where Exp Replay totally fails. This exciting result demonstrates that the OWL agent is able to re-use each of the base policies learnt sequentially for solving totally different tasks. Randomly selecting policies in the roll-out is a strong strategy which the MAB can emulate. This demonstrates potential for our approach in a hierarchical RL setting, with links to options \cite{optioncritic}.

\subsection{Ablations}
\label{sec:ablations}
OWL decreases in performance when we remove the EWC regularization which helps to prevent forgetting in Table~\ref{tab:minigrid_ablations_main_paper}. By replacing the EWC regularization with a distillation loss which ensures that the outputs from the previous task's $Q$-function head remain similar to the previous task's $Q$-function head for the current task \cite{hinton2015distilling, Li2017}, also decreases performance, see Section~\ref{sec:minigrid_ablations} for more details. These regularizations works well with classification problems, however we are performing a regression, this difference could help explain the drop in performance with EWC. We also compare to a Full Rehearsal (FR) which is an upper bound to OWL performance. FR has a buffer for each task and a separate policy head for each task, as such it does not scale gracefully as the number of tasks increase in comparison to OWL, see Section~\ref{sec:full_rehearsal} in the appendix for implementation details.

{\centering
\begin{table}
\vspace{-0mm}
\centering
\caption{\small{Comparisons and ablations for OWL evaluating on the $5$ SC and SC+DK tasks for $10$ seeds.}}
\label{tab:minigrid_ablations_main_paper}
\scalebox{1.0}{
\begin{tabular}{lcc}
    \toprule
      & \textbf{SC} & \textbf{SC+DK} \\
      \midrule
     Exp Replay & $0.01$ \scalebox{0.75}{$(0.61, 0.00)$} & $0.00$ \scalebox{0.75}{$(0.52, 0.00)$}\\
     OWL (orcl) & $\textbf{0.85}$ \scalebox{0.75}{\textbf{$(0.97, 0.72)$}} & $\textbf{0.60}$ \scalebox{0.75}{$\textbf{(0.98, 0.44)}$} \\
     OWL (adpt) & $0.59$ \scalebox{0.75}{$(0.75, 0.48)$} & $\textbf{0.63}$ \scalebox{0.75}{$\textbf{(0.79, 0.45)}$} \\
     OWL - EWC (orcl) & $0.45$ \scalebox{0.75}{$(0.53, 0.39)$} & $0.40$ \scalebox{0.75}{$(0.48, 0.30)$} \\
     OWL - EWC (adpt) & $0.49$ \scalebox{0.75}{$(0.60, 0.39)$} & $\textbf{0.50}$ \scalebox{0.75}{$\textbf{(0.62, 0.37)}$} \\
     OWL - EWC + DL (orcl) & $0.45$ \scalebox{0.75}{$(0.53, 0.36)$} & $0.34$ \scalebox{0.75}{$(0.40, 0.29)$} \\
     OWL - EWC + DL (bndt) & $0.53$ \scalebox{0.75}{$(0.61, 0.38)$} & $0.39$ \scalebox{0.75}{$(0.45, 0.33)$} \\
     \midrule
     Full Rehearsal & $0.99$ \scalebox{0.75}{$(0.99, 0.97)$} & $0.99$ \scalebox{0.75}{$(1.00, 0.98)$} \\
     \bottomrule
    \end{tabular}}
    \vspace{-6mm}
\end{table} 
}

\section{Conclusion and Future Work}
\label{sec:conclusion}

In this paper we consider a challenging continual reinforcement learning setting where \emph{different tasks} have the \emph{same observation}. We showed that established experience replay methods which are task agnostic with a single predictor network fail due to interference. Our main contribution is to highlight this interference problem and introduce a simple yet effective approach for this paradigm, which we call OWL. OWL is able to limit forgetting while training on  tasks sequentially by using an $Q$-function with a shared feature extractor and a population of linear heads for each task. OWL does not require knowledge of the task at test time, but is still able to achieve close to optimal performance using an algorithm inspired by multi-armed bandits. We evaluated OWL on challenging RL environments such as MiniGrid, where we were able to solve five different tasks with similar observations. Finally, we showed it is possible to transfer our learned policies to entirely unseen and more challenging environments.

There are a variety of exciting future directions for this work. For instance, it is desirable to detect task boundaries during training in addition to evaluation. It would be interesting to explore change detection methods and have more robust probabilistic models in RL which are able to detect shifts in reward distributions and state-action distributions of new tasks to enable learning new tasks continually. In the MiniGrid experiments it would be interesting to learn a way to learn a curriculum of environments such that they induce a set of policies with the skills/behaviors required to generalize to even harder tasks. 

\appendix
\onecolumn

\section*{Appendices}

\section{Limitations}
\label{sec:limitations}
In Q-learning it is the ranking of the Q-values conditioned on a state that is important rather than the accuracy of the Q-values themselves; indeed we observe empirically that TD error increases with training as we encounter higher return episodes. With very diverse tasks (e.g., sparse \emph{and} dense reward) this could result in a diverse range of TD errors which will require normalization when trying to facilitate MAB policy selection; for instance policy heads corresponding to sparser reward tasks may be over-selected simply because they have intrinsically lower TD error. We have chosen similar tasks to demonstrate interference where TD error is comparable across tasks, but accept that tasks with diverse TD error may need a different feedback to the MAB.

\begin{figure}
    \centering
    \includegraphics[width=0.4\linewidth]{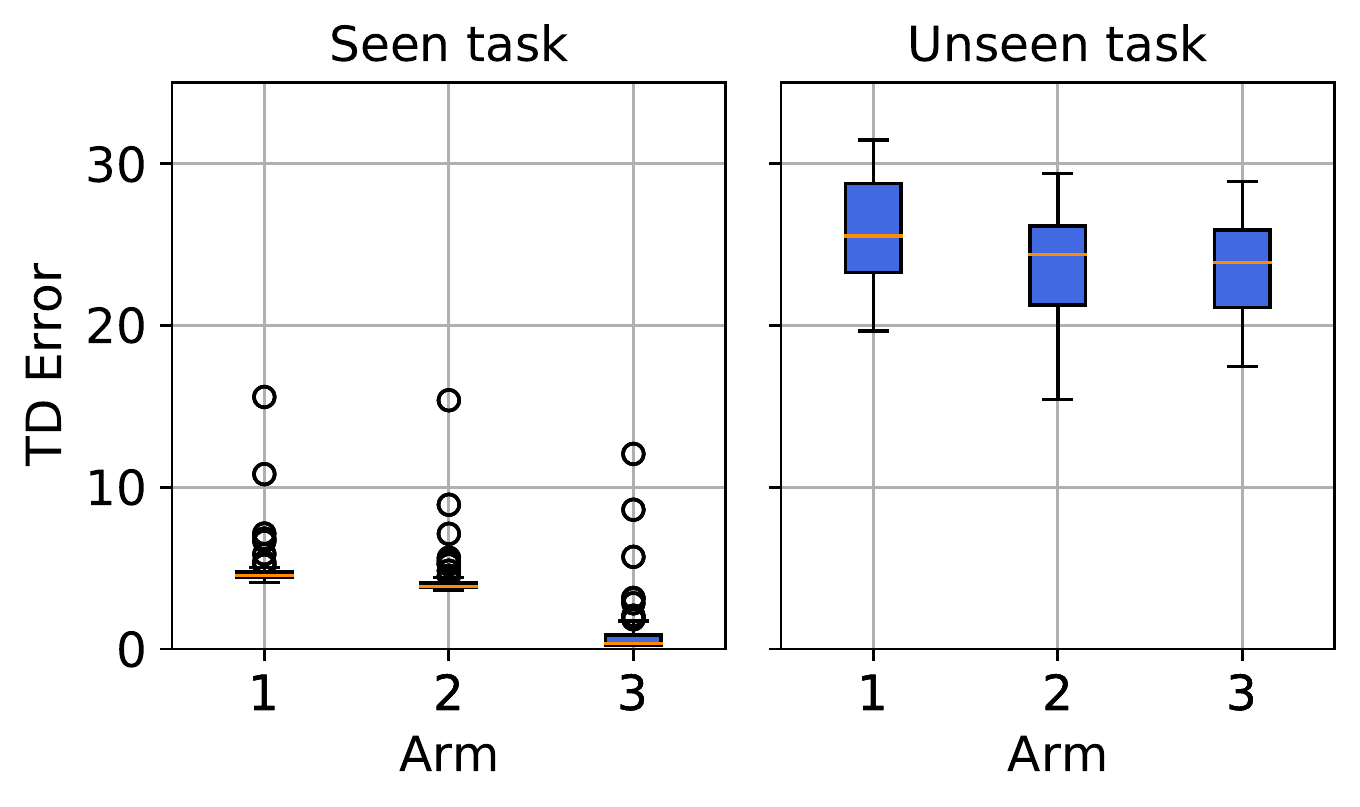} 
    \caption{\small{TD errors for each head of a $3$ head/policy OWL agent for a task which OWL has seen before \textbf{Left} task 3 in Figure~\ref{fig:minigrid_t3_ewc500}. \textbf{Right}, TD errors for an unseen $5$ walled $\textrm{SimpleCrossing}$ task.}}
    \label{fig:td_error_in_vs_out}
\end{figure}

There is a statistically significant benefit to using the MAB for in distribution tasks which have been seen during continual training. However the benefits do not translate for tasks which are unseen. The default behaviour of the MAB is to behave randomly and assign a uniform distribution/prior on all policy heads; this is intended. It seems that using DQN as a base learner for $5$ tasks is not enough for our policy to truly generalize to the distribution of tasks. Thus the feedback to MAB is ensuring that MAB probabilities are uniform. This is shown in Figure~\ref{fig:td_error_in_vs_out}, on the left we show the TD errors of the seen task where the TD error for arm 3 is the lowest; the MAB is correctly picking the arm/policy to solve the task. On the right, an unseen $5$ wall task where all TD errors are high: there isn't a signal for the MAB to latch onto. There isn't enough generalization in the 3 polices of this agent for the agent to assign a non-uniform distribution for the MAB arms/policies hence the random behaviour of the MAB. If we could endow DQN with invariances through for instance self-supervision such that the policies can generalise then the feedback to the MAB would be stronger and outperform randomly picking the policy head.

We have on purpose picked the Minigrid sparse reward environments as they are difficult to solve. If the MAB has a dense reward like in the Pendulum then it can quite easily settle on the correct task just by using the reward as feedback.

A strength of OWL is its simplicity. However, there is still a performance gap between the optimal performance achievable: FR versus OWL. There is still some forgetting as we scale to more tasks and train for less time. Thus the effectiveness of EWC as a method to prevent forgetting is a further limitation. Our paper focuses on preventing interference and task inference at test time but the use of stronger methods to prevent forgetting is an interesting extension.

\section{MultiMNIST Experiments}

In terms of dataset construction: two different classes from the MNIST dataset are placed on the top left and bottom right of each image and each randomly shifted $4$ pixels in each direction. The resulting images are of size $64 \times 64$. The first task $\mathcal{T}_1$ requires classifying the top right image and the second task $\mathcal{T}_2$ requires classifying the bottom right: these two tasks are conflicting. We use EWC and GEM continual learning strategies \cite{ewc, Lopez-Paz} with LeNet backbones.

\section{Implementation Details}
\label{app:implmentation_details}
In this section we detail the implementation of OWL applied for SAC which was used in the continuous control setting with the $\mathrm{Pendulum}$-$\mathrm{v0}$ environments. Also we detail the implementation of OWL applied to DQN for solving the MiniGrid environments.

All experiments where run on a single NVIDIA RTX 3090 GPU, with multiple seeds fitting onto the GPU.

\subsection{Conflicting Pendulums: Soft Actor-Critic}
\label{sec:sac_implementation_details}

We implement Soft Actor-Critic (SAC) in PyTorch, following the learned reward temperature approach of \cite{sac-v2}. Briefly, Soft Actor-Critic aims to maximize the sum of the reward and the entropy of the policy over the task horizon; this results in behaviour that can be summarized as ``maximising reward while acting as randomly as possible", and can be shown as being optimal in a meta-POMDP setting \cite{eysenbach2019maxent}. What is difficult about such a dual objective is that ultimate task performance is sensitive to the trade-off between randomness/entropy and reward \cite{sac}. In the continual setting that we present here, different tasks over the agent's lifetime may require different degrees of reward/entropy scaling, which would require specific tuning of this parameter (usually denoted as $\alpha$) per new task. It is possible however to learn this parameter $\alpha$ from the task by introducing a `target entropy' ($\bar{\mathcal{H}}$), which actively scales reward against policy entropy such that the expected entropy matches the target $\bar{\mathcal{H}} = -\text{dim}(\mathcal{A})$, where $\mathcal{A}$ is the action space of an environment. This results in $\alpha$ that adapts to its current policy and experiences it receives from the environment, as well as an appropriate degree of randomness for each environment.

For Exp Replay and SAC we use a buffer of size $1$M, OWL uses an EWC regularization strength of $\lambda=100$. In this setting we use an ensemble of probabilistic networks and thus we are able to provide a negative log-likelihood of the Q-values for each policy as feedback to the bandit algorithm $l^t_{i_t}$ for deciding which Q-function and policy to use to solve the task which is being evaluated on. 

The Q-function used for OWL is identical to the architecture from PETS which has been shown to be successful in Model-Based RL \cite{pets}. OWL uses a multi-head architecture where the number of heads equals the number of tasks. Each head has a mean and variance mapping \cite{weigend}. SAC uses Q-functions with four layers with ReLU activations. An ensemble of two Q-functions are used for compensation for an optimistic Q-value as is standard for SAC. The target Q-function $Q(\cdot, \cdot;\theta_{i-1})$ is updated using a exponentially weighted moving average with $\gamma = 0.99$.

The Gaussian policy is parameterized by a NN which uses mean and variance heads and share three NN layers with ReLU activations, it is trained with a learning rate of $3 \times 10^{-4}$. Gradients are propagated through the Gaussian policy using the reparameterization trick \cite{Kingma2013AutoEncodingVB}.

\subsubsection{Elastic Weight Consolidation}

The EWC implementation for both the Q-function and policy uses $\lambda = 100$, we didn't attempt to optimize $\lambda$. The weighting of the $L^2$ regularization is performed with the empirical Fisher Information $F(\theta) = \sum_n \nabla_{\theta} \log p_{\theta}(y_n|x_n) \nabla_{\theta} \log p_{\theta}(y_n|x_n)^{\top}$ which is calculated using all samples from the experience replay buffer upto a maximum of $60,000$ samples. EWC is applied on the feature extraction layers of the Q-function and policy and to the task specific mean and variance heads for when they are trained upon again. No EWC regularization is directly applied to the target Q-function.

\subsubsection{Online Learning}
\label{sec:ol_deets}

Two Q-values are estimated by SAC and OWL. In OWL the Q-function outputs a mean and variance, these estimates can be combined by considering the Q-values as a mixture of Gaussians. The Gaussian mixture of $L$  Q-functions has mean and variance $\mu_{*} = \frac{1}{L} \sum^{L}_{l=1}\mu_{l}(x)$ and $\sigma^2_{*} = \sum^{L}_{l=1} (\sigma^2_{l}(x) - \mu^{2}_{l}(x)) - \mu^{2}_{*}(x)$ \cite{Lakshminarayanan}.  These estimates are then used as the basis for MSE and negative log-likelihood feedback to the Exponentially Weighted Average Forecaster bandit algorithm (ExpWeights). More specifically the feedback is $G_{\phi_i}(\theta_i) = \log \sigma_{*}^{2}(s_t, a_t;\theta_i) + \frac{(\mu_{*}(s_t, a_t;\theta_i)) - \hat{\mu}_{*}(s_t, a_t;\theta_i)}{\sigma_{*}^{2}(s_t, a_t;\theta_i)}$ for head $\theta_i$ and $\hat{\mu}(\cdot)$ is the target mean Q-function, the feedback to the MAB is thus $l_{i_t} = 1/G_{\phi_i}(\theta_i)$.

The losses supplied to ExpWeights are sometimes very large, especially for a feedback in the form of a MSE. Hence we set a threshold to $l_{i_t} = \textrm{min}(50, l_{i_t})$ where $l$ is a loss, those considered are inverse negative log-likelihood and inverse MSE, $i_t$ denotes the index of the policy chosen at time $t$ in the rollout. The step size is set to $\eta = 0.98$. We haven't attempted to optimize these. The MAB algorithm which we use for OWL is summarized in Algorithm~\ref{alg:unclear_test}.

\subsection{Minigrid: DQN}
\label{sec:dqn_implementation_details}
We use a DQN implementation to solve the MiniGrid environments and adapt our OWL method accordingly. For the Q-function we use Duelling Networks \cite{Wang2016} and use Double-DQN for estimating the Q-value for a state-action \cite{VanHasselt2015}. For the Minigrid environment the inputs are images of size $7 \times 7 \times 3$. We focus on tasks generated from the $\mathrm{SimpleCrossingS9N1}$ environment. We use a  CNN to as a function approximator for learning Q-values, see Table~\ref{tab:dqn_cnn_arch} for a description of the architecture used. To enable exploration of the Minigrid state-space we use an $\epsilon$-greedy exploration strategy which is annealed over the course of training. In addition, we use a count based reward bonus for visiting different places on the grid and if the agent reaches the goal during training the reward is multiplied by $100$ to boost the signal and differentiate the reward signal from the state based exploration bonus. We limit the action space to $\mathcal{A} = \{\textrm{left}, \textrm{right}, \textrm{forward}\}$ for the SC agent.

We use an experience replay buffer of size $1$M frames which is standard in DQN implementations. Crucially this buffer is emptied when learning a new task with OWL. For our Exp. Replay baseline experiences persist across different tasks. The target Q-function network is assigned the parameters from the Q-function every $80$ optimization steps. The Adam optimizer \cite{Kingma2015} is used for the Q-function with a Huber loss. See Table~\ref{tab:dqn_hparams} for the list of hyperparameters used and their values, most values are set to best practice values \cite{Deepmind2017}.

\subsubsection{OWL}

To enable learning multiple tasks sequentially without interference we use a shared feature extractor and a different linear head per task. Additionally, we clear the experience replay buffer before learning a new task so that experiences do not interfere. To alleviate forgetting we use EWC regularization of the shared feature extractor. We found $\lambda=500$ worked well, see Section~\ref{sec:ablations}.

\subsubsection{Online Learning} We compute the feedback to the bandit algorithm similarly to the $\mathrm{Pendulum}$-$\mathrm{v0}$ experiment. However we use the TD error as feedback $G_{\phi_i}(\theta_i) = (Q(s_t, a_t; \theta_i) - \hat{Q}(s_t, a_t; \theta_i))^2$ for Q-function head $\theta_i$, $\phi_i$ is the policy head which is the same as $\theta_i$ in DQN since there is no separate policy like in SAC and $\hat{Q}(\cdot)$ is the target. We found that the TD error worked well as feedback. We optimized the bandit algorithm's hyperparameters and found $\eta=0.88$ worked well.

\begin{table}
\centering
    \caption{DQN hyperparameters used for Exp. Replay and OWL methods.}
     \label{tab:dqn_hparams}
     \begin{tabular}{lc}
    \toprule
      Parameter & Value \\ 
      \midrule
     Min number of random experiences to start learning & $10$K frames\\
     Discount & $0.99$ \\
     Adam learning rate & 0.0000625 \\
     Batch size & $32$ \\
     Start exploration $\epsilon$ & 0.9\\
     Min exploration $\epsilon$ & 0.01\\
     Exploration decay rate & $250$k steps \\
     Target network update frequency & Every $80$ opt. steps \\
     Q-function update frequency & Every $4$ env. steps \\
     Experience replay size & $1$M frames\\
     Maximum evaluation steps in env. & $100$ \\ 
     Number of episodes for evaluation & $16$ \\
     Frames per task & $1$M \\
     \bottomrule
    \end{tabular}
\end{table}

\begin{table}
\centering
    \caption{DQN Q-function architecture used for Exp. Replay and OWL methods.}
     \label{tab:dqn_cnn_arch}
     \begin{tabular}{lcccc}
    \toprule
      Layer & Channel & Kernel & Stride & Padding \\ 
      \midrule
      Input $7 \times 7$ & 3 & - &-  &-  \\
      Conv 1 & 16 & ($2\times2$) & 1 & 0 \\
      ReLU  & - & - & - & -  \\
      Max Pool $2$-d & 16 & ($2\times2$) & 2 & 0 \\
      Conv 2 & 32 & ($2\times2$) & 1 & 0 \\
      ReLU  & - & - & - & -  \\
      Conv 3 & 64 & ($2\times2$) & 1 & 0 \\
      ReLU  & - & - & - & -  \\
      Flatten  & - & - & - & -  \\
      Linear  & $200$ & - & - & -  \\
      ReLU  & - & - & - & -  \\
      \midrule
      Value Stream & $1$ & - & - & - \\
      Advantage Stream & $|\mathcal{A}|$ & - & - & - \\
     \bottomrule
    \end{tabular}

\end{table}

\section{Ablation Studies}
\label{app:ablations}

\subsection{Conflicting Pendulums}

\begin{figure}[h]
    \includegraphics[width=.50\linewidth]{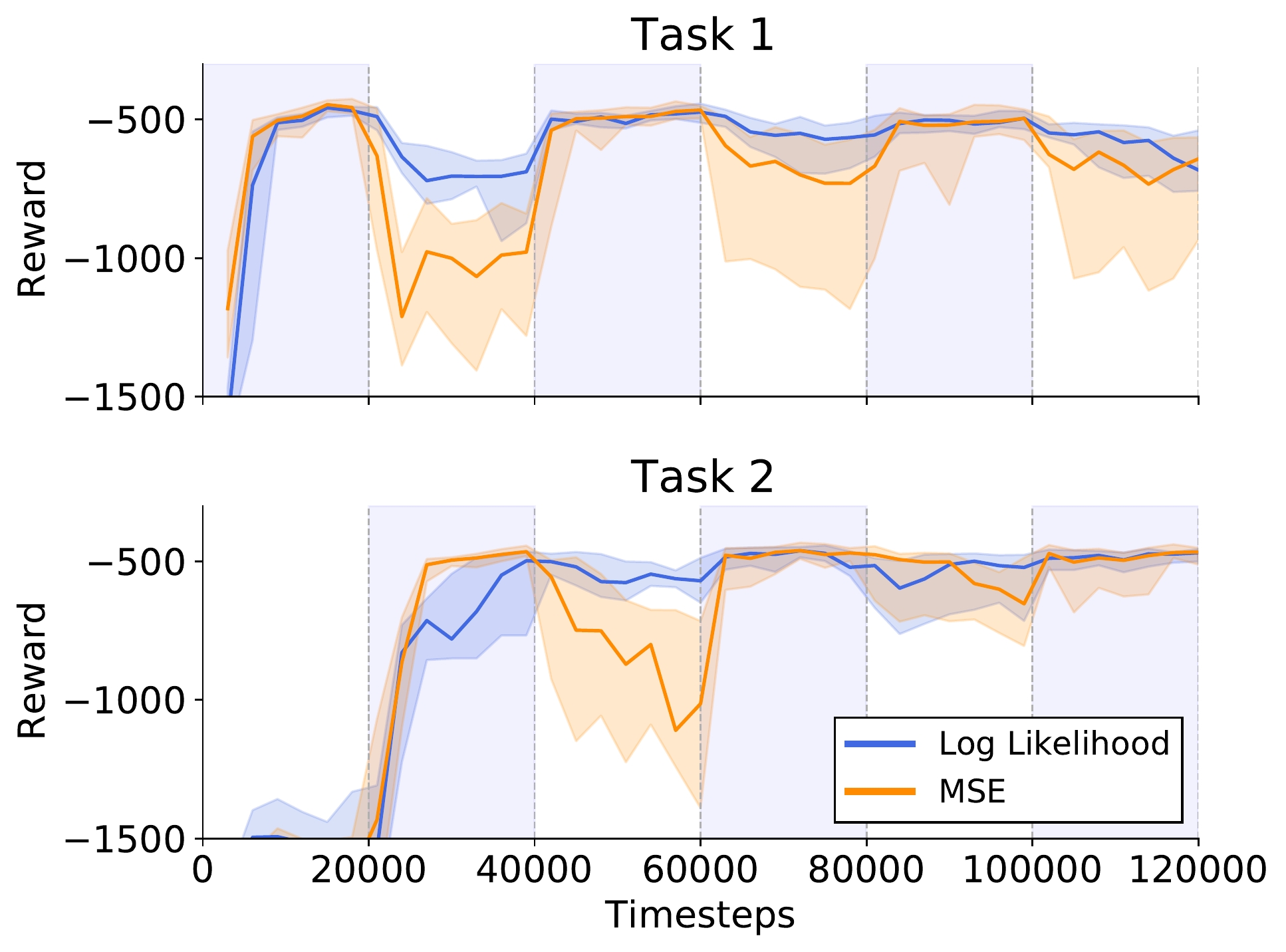} 
    \includegraphics[width=.48\linewidth]{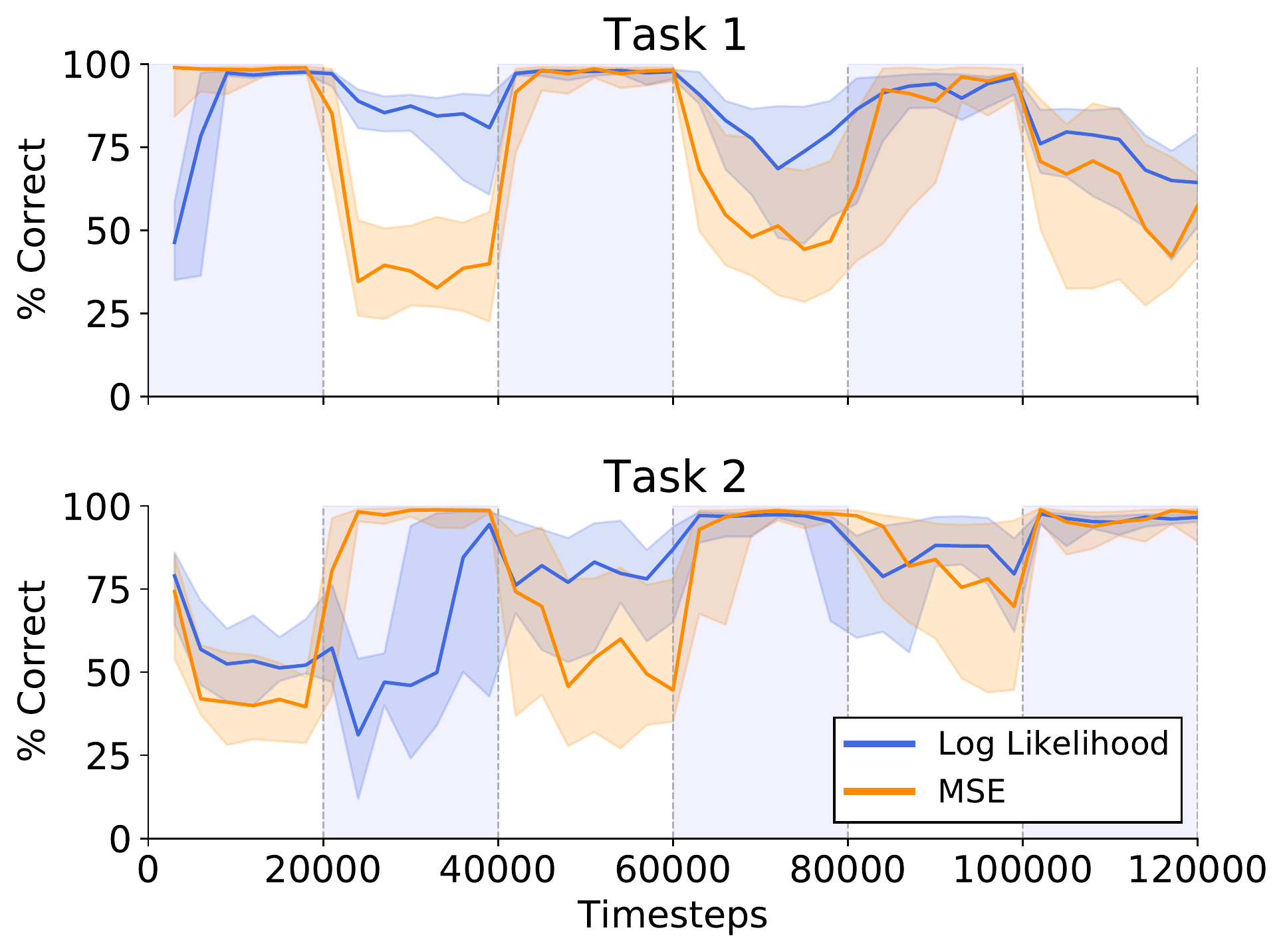} 
    \vspace{-3mm}
    \caption{Left: Median performance across $10$ seeds. Shaded region enveloping the lines correspond to the inter-quartile range. Right: The performance of the online learning mechanism for both tasks. Pale blue alternating tiles correspond to the task which is currently being trained on.}
    \label{figure:ablation}
\end{figure}

\paragraph{Do we need uncertainty aware models?} The feedback for our online learning mechanism for OWL adapted to SAC incorporates both the mean and variance predictions, thus considering aleatoric uncertainty. To assess the impact of this, we also ran OWL with a simple MSE feedback, ignoring the variance head. In Fig. \ref{figure:ablation} we show both the performance of this approach, as well as the effectiveness of the online learning mechanism.

It is clear to see here that the negative log-likelihood is significantly more effective. The linkage between selecting the correct policy and final performance is evident, and justifies our use of the probabilistic models. 

\subsection{Minigrid}
\label{sec:minigrid_ablations}

We analyse the performance of the different OWL variants by look at performance on $2$ Minigrid tasks repeated twice.

\textbf{EWC regularization strength:} We can vary the strength of the EWC regularization, $\lambda$. The larger $\lambda$ the less forgetting of previous tasks, however the additional regularizations can inhibit learning new tasks. We do not want $\lambda$ to be too large or too small. We found $\lambda=500$ to work well, in general performance isn't very sensitive to $\lambda$, see Table~\ref{tab:minigrid_ewc_ablation}.

\textbf{Functional regularization:} Different parameterizations of a NN can yield the same output function. Instead of regularizing the weights of a NN it is more desirable to regularize the output function of a NN such that past outputs look same as current outputs \cite{hinton2015distilling,Li2017, Benjamin2019}. Since we use a shared feature extractor $f_{\theta_z}$ and different outputs heads $\theta_{1:M}$ to parameterize the Q-function. At the end of training on $\mathcal{T}_{\tau}$ we can cache the parameters of the model and use this old model to regularize the new model. While training the new task we can ensure that the shared feature extractor's outputs do not deviate from old values by regularizing the outputs with respect to the previous task's outputs. The functional regularization for task $\mathcal{T}_{\tau+1}$ is
\begin{align}
\label{eq:func_reg}
\mathcal{L}^{\tau+1}_{\textrm{func}} 
&= \mu \, \mathrm{KL}(Q_{\tau+1}\left((\cdot, \cdot), Q_{\tau}(\cdot, \cdot) \right)\\
&= \mu \left(g_{\theta_{\tau}} \circ f_{\theta_{z, \tau+1}} (\cdot, \cdot)  - g_{\theta_{\tau}} \circ f_{\theta_{z, \tau}} (\cdot, \cdot) \right)^2. 
\end{align}
Where we assume that the variances of the Q-functions are equal and constant. $\theta_{\tau}$ are the output head parameters of the previous task and $g_{\theta_{\tau}}$ is the linear head, $\circ$ denotes a composition of functions. $\theta_{z, \tau+1}$ are the parameters of the shared feature extractor for the current task $\mathcal{T}_{\tau+1}$ and $\mu$ is a hyperparameter which controls the strength of the regularization. On $2$ Minigrid tasks repeated twice we found that the functional regularization doesn't perform as well as EWC regularization for different values of the regularization strength $\mu \in \{0.1, 1, 10, 100\}$. The best value of $\mu$ we found was $100$.

\begin{table}
\centering
    \caption{Median proportion of successes for $5$ seeds after $4$M frames for $2$ Minigrid tasks ($75$-th quartile, $25$-th quartile). Performance is stable to different EWC regularization strengths, $\lambda$.}
     \label{tab:minigrid_ewc_ablation}
     \resizebox{\textwidth}{!}{\begin{tabular}{lccccc}
    \toprule
      & \multicolumn{2}{c}{Oracle} & \multicolumn{2}{c}{Adaptive} &\\ 
      \midrule
      & Final Perf. & Cumulative Perf. & Final Perf. & Cumulative Perf. & Avg. Rank \\
      \midrule
     Exp Replay & $0.00$ \scalebox{0.75}{$(0.09, 0.00)$}
     & $0.38$ \scalebox{0.75}{$(0.87, 0.01)$} & - & - & 4.0 \\
     \midrule
      OWL $\lambda=200$ & $1.00$ \scalebox{0.75}{$(1.00, 1.00)$} & $0.84$ \scalebox{0.75}{$(0.91, 0.76)$} & $1.00$ \scalebox{0.75}{$(1.00, 0.79)$} & $0.73$ \scalebox{0.75}{$(0.85, 0.61)$} & 2.0\\
      OWL $\lambda=500$ & $1.00$ \scalebox{0.75}{$(1.00, 1.00)$} & $0.92$ \scalebox{0.75}{$(0.95, 0.89)$} & $1.00$ \scalebox{0.75}{$(1.00, 0.64)$} & $0.75$ \scalebox{0.75}{$(0.88, 0.61)$} & 1.25\\
      OWL $\lambda=1000$ & $1.00$ \scalebox{0.75}{$(1.00, 1.00)$} & $0.91$ \scalebox{0.75}{$(0.92, 0.75)$} & $1.00$ \scalebox{0.75}{$(1.00, 1.00)$} & $0.86$ \scalebox{0.75}{$(0.89, 0.84)$} & 1.25\\
     \bottomrule
    \end{tabular}}
\end{table}

\textbf{Warm starting the replay buffer:} DQN initializes the experience replay buffer by drawing $10^5$ random actions before using the DQN policy to select actions for the first task. In OWL we flush the experience replay buffer when starting to train in a new task to avoid interference, however this means that our experience replay buffer is empty for a new task and we would start training with no experiences. Alternatively we could sample some experiences from each task and use these to warm start the experience replay buffer when revisiting these tasks. We found this increased the performance of OWL when warm starting the experience replay buffer with $B = \{5\times 10^4, 10^5\}$ experiences, see Table~\ref{tab:minigrid_ablations}. Our results in Table~\ref{tab:minigrid_main_results} include this technique with $B=5 \times 10^4$.

\textbf{Restarting the $\epsilon$ greedy exploration strategy:} We keep a separate $\epsilon$-greedy exploration schedule for each task for OWL. When we revisit a task we simply look up the schedule we were using the last time we were training in the task and continue where the $\epsilon$-greedy schedule left off, this approach is used in \cite{ewc}. However, when revisiting a task it makes sense to add a bit of exploration as a warm start for the experience replay buffer, random actions are added to the experience replay buffer when training on DQN. We try resetting the $\epsilon$-greedy schedule when revisiting a task and annealing at the same rate ($1 \times \tau_{\epsilon}$) or annealing $\epsilon$ at twice the rate when revisiting the task ($2 \times \tau_{\epsilon}$). We found poor performance for both these scenarios. It is best to keep an $\epsilon$-greedy schedule for each task pick the up the schedule where the DQN agent left off when revisiting a task, see Table~\ref{tab:minigrid_ablations}.

\begin{table}
\centering
    \caption{Median proportion of successes for $5$ seeds after $4$M frames and cumulative performance over $4$M frames and ($75$-th quartile, $25$-th quartile). All ablations have EWC regularization with $\lambda=500$.}
     \label{tab:minigrid_ablations}
     \resizebox{\textwidth}{!}{\begin{tabular}{lccccc}
    \toprule
     & \multicolumn{2}{c}{Oracle} & \multicolumn{2}{c}{Adaptive} & \\ 
      \midrule
      & Final Perf. & Cumulative Perf. & Final Perf. & Cumulative Perf. & Avg. Rank \\
      \midrule
     Exp Replay & $0.00$ \scalebox{0.75}{$(0.09, 0.00)$} & $0.38$ \scalebox{0.75}{$(0.87, 0.01)$} & - & - & 6.0\\
     \midrule
     OWL warm start $50$k & $1.00$ \scalebox{0.75}{$(1.00, 1.00)$} & $0.92$ \scalebox{0.75}{$(0.93, 0.90)$} & $1.00$ \scalebox{0.75}{$(1.00, 0.88)$} & $0.88$ \scalebox{0.75}{$(0.89, 0.81)$} & 1.0 \\
     OWL warm start $100$k & $1.00$ \scalebox{0.75}{$(1.00, 1.00)$} & $0.92$ \scalebox{0.75}{$(0.95, 0.87)$} & $1.00$ \scalebox{0.75}{$(1.00, 0.88)$} & $0.81$ \scalebox{0.75}{$(0.87, 0.67)$} & 1.5 \\
     OWL $\epsilon$-greedy warm start $1 \times \tau_{\epsilon}$ & $1.00$ \scalebox{0.75}{$(1.00, 1.00)$} & $0.78$ \scalebox{0.75}{$(0.85, 0.68)$} & $0.97$ \scalebox{0.75}{$(1.00, 0.80)$} & $0.71$ \scalebox{0.75}{$(0.75, 0.64)$} & 3.5\\
     OWL $\epsilon$-greedy warm start $2 \times \tau_{\epsilon}$ & $1.00$ \scalebox{0.75}{$(1.00, 1.00)$} & $0.82$ \scalebox{0.75}{$(0.87, 0.69)$} & $0.80$ \scalebox{0.75}{$(0.98, 0.62)$} & $0.64$ \scalebox{0.75}{$(0.78, 0.60)$} & 3.5 \\
     OWL MLP head & $1.00$ \scalebox{0.75}{$(1.00, 1.00)$} & $0.88$ \scalebox{0.75}{$(0.92, 0.88)$} & $1.00$ \scalebox{0.75}{$(1.00, 1.00)$} & $0.85$ \scalebox{0.75}{$(0.89, 0.75)$} & 1.75 \\
     \bottomrule
    \end{tabular}}
\end{table}

\textbf{MLP head:} Multi-head models are a very useful tool and widely used in CL \cite{vcl} a feature extractor $z = f_{\theta_z}(x)$ is shared for all tasks and a task specific linear head $g_{\theta_i}(z)$ is appended for each individual task. We can add additional flexibility by making the head $g_{\theta_i}(z)$ a $2$ layer MLP. This wasn't shown to help, Table~\ref{tab:minigrid_ablations}.

\textbf{MC dropout:} Can we obtain uncertainties for our Q-function and can this be a better guide / feedback for the bandit algorithm? We showed that providing a negative log-likelihood feedback improved results for SAC in $\mathrm{Pendulum}$-$\mathrm{v0}$. Can we obtain uncertainties for DQN? Having a probabilistic network with a variance head didn't work; the agent wasn't able to learn anything by optimizing a negative log-likelihood as an objective function instead of a Huber loss. We turn our attention to using MC dropout which is also very easy to implement for our current set up: we need to 1. train our Q-function with dropout and at evaluation 2. use dropout to make predictions \cite{Gal2016}. We found that performance degrades considerably when enabling dropout at different rates, Table~\ref{tab:minigrid_mc_dropout_ablation}. The Oracle performance decreases, hence training with dropout provides poor policies. Additionally using the resulting negative log-likelihood of the TD error as feedback also doesn't help performance of the bandit algorithm. The main results  (Table~\ref{tab:minigrid_main_results}) already show that using a TD error for feedback to the bandit algorithm is sufficient feedback for OWL solve for an environment.

\begin{table}
\centering
    \caption{Median proportion of successes for $5$ seeds after $4$M frames and cumulative performance over $4$M frames of $2$ different MiniGrid tasks repeated twice ($75$-th quartile, $25$-th quartile). Enabling dropout regularization and using a negative log-likelihood feedback for the bandit algorithm hurts performance. Average ranks computed for warm-start and no warm start OWL methods separately with Experience Replay.}
     \label{tab:minigrid_mc_dropout_ablation}
    \resizebox{\textwidth}{!}{\begin{tabular}{lccccc}
    \toprule
      & \multicolumn{2}{c}{Oracle} & \multicolumn{2}{c}{Adaptive} &\\ 
      \midrule
      & Final Perf. & Cumulative Perf. & Final Perf. & Cumulative Perf. & Avg. Rank \\
      \midrule
     Exp Replay & $0.00$ \scalebox{0.75}{$(0.09, 0.00)$}
     & $0.38$ \scalebox{0.75}{$(0.87, 0.01)$} & - & - & 4.0 \\
     \midrule
      OWL $\lambda=500$ & $1.00$ \scalebox{0.75}{$(1.00, 1.00)$} & $0.92$ \scalebox{0.75}{$(0.95, 0.89)$} & $1.00$ \scalebox{0.75}{$(1.00, 0.64)$} & $0.75$ \scalebox{0.75}{$(0.88, 0.61)$} & 1.25\\
      OWL $d=0.05$ & $1.00$ \scalebox{0.75}{$(1.00, 1.00)$} & $0.85$ \scalebox{0.75}{$(0.93, 0.61)$} & $1.00$ \scalebox{0.75}{$(1.00, 0.87)$} & $0.66$ \scalebox{0.75}{$(0.74, 0.60)$} & 1.75\\
      OWL $d=0.1$ & $1.00$ \scalebox{0.75}{$(1.00, 0.25)$} & $0.69$ \scalebox{0.75}{$(0.76, 0.52)$} & $0.88$ \scalebox{0.75}{$(0.99, 0.20)$} & $0.59$ \scalebox{0.75}{$(0.74, 0.52)$} & 2.75\\
      \midrule
      OWL warm start $50$k $d=0.0$ & $1.00$ \scalebox{0.75}{$(1.00, 1.00)$} & $0.92$ \scalebox{0.75}{$(0.93, 0.90)$} & $1.00$ \scalebox{0.75}{$(1.00, 0.88)$} & $0.88$ \scalebox{0.75}{$(0.89, 0.81)$} & 1.0 \\
      OWL warm start $50$k $d=0.05$ & $1.00$ \scalebox{0.75}{$(1.00, 1.00)$} & $0.88$ \scalebox{0.75}{$(0.93, 0.63)$} & $1.00$ \scalebox{0.75}{$(1.00, 0.88)$} & $0.70$ \scalebox{0.75}{$(0.82, 0.54)$} & 1.5\\
      OWL warm start $50$k $d=0.1$ & $1.00$ \scalebox{0.75}{$(1.00, 1.00)$} & $0.80$ \scalebox{0.75}{$(0.85, 0.78)$} & $0.95$ \scalebox{0.75}{$(1.00, 0.40)$} & $0.73$ \scalebox{0.75}{$(0.80, 0.59)$} & 2.25\\
      OWL warm start $50$k $d=0.2$ & $0.99$ \scalebox{0.75}{$(1.00, 0.24)$} & $0.46$ \scalebox{0.75}{$(0.62, 0.29)$} & $0.77$ \scalebox{0.75}{$(0.96, 0.17)$} & $0.44$ \scalebox{0.75}{$(0.57, 0.29)$} & 3.75\\
     \bottomrule
    \end{tabular}}
\end{table}

\section{Scaling to $3$ $\textrm{SimpleCrossing}$ tasks.}

Above, we considered a variety of design choices for OWL. In particular, we found that warm starting the experience replay buffer, denoted as OWL+WS in Table~\ref{tab:minigrid_main_results} with a small sample of experiences from the same task upon revisiting it worked well. Finally, we used $\lambda = 500$ for the EWC regularization strength. The results on $3$ Minigrid levels which accompany the learning curves in Figure~\ref{fig:minigrid_t3_ewc500} are summarized in Table~\ref{tab:minigrid_main_results}.

\begin{table}
\centering
\caption{Median proportion of successes for three different MiniGrid environments (tasks) and inter-quartile range for $10$ seeds after $9$M frames and $3$ task repeats and cumulative performance over the same period. Our methods are able to solve all three tasks, without knowledge of the test task, while Experience Replay fails due to interference.}
\vspace{+0.2cm}
\label{tab:minigrid_main_results}
\scalebox{1.0}{
\begin{tabular}{lcc}
    \toprule
      & \textbf{Final Perf.} & \textbf{Cumulative Perf.} \\
      \midrule
     Exp Replay & $0.00$ \scalebox{0.75}{$(1.00, 0.00)$}
     & $0.37$ \scalebox{0.75}{$(0.49, 0.12)$} \\
      OWL (Orcl.) & $1.00$ \scalebox{0.75}{$(1.00, 1.00)$} & $0.83$ \scalebox{0.75}{$(0.94, 0.71)$} \\
      OWL + WS (Orcl.) & $1.00$ \scalebox{0.75}{$(1.00, 1.00)$} & $0.95$ \scalebox{0.75}{$(0.97, 0.89)$} \\
      OWL (Adpt.) & $0.86$ \scalebox{0.75}{$(1.00, 0.70)$} & $0.72$ \scalebox{0.75}{$(0.77, 0.62)$} \\
      OWL + WS (Adpt.) & $0.99$ \scalebox{0.75}{$(1.00, 0.83)$} & $0.79$ \scalebox{0.75}{$(0.91, 0.73)$} \\
     \bottomrule
    \end{tabular}}
\end{table}

\section{Scaling to $5$ Minigrid tasks}

\begin{figure*}[h]
    \hspace*{-1cm}
    \centering
    \includegraphics[width=0.95\linewidth]{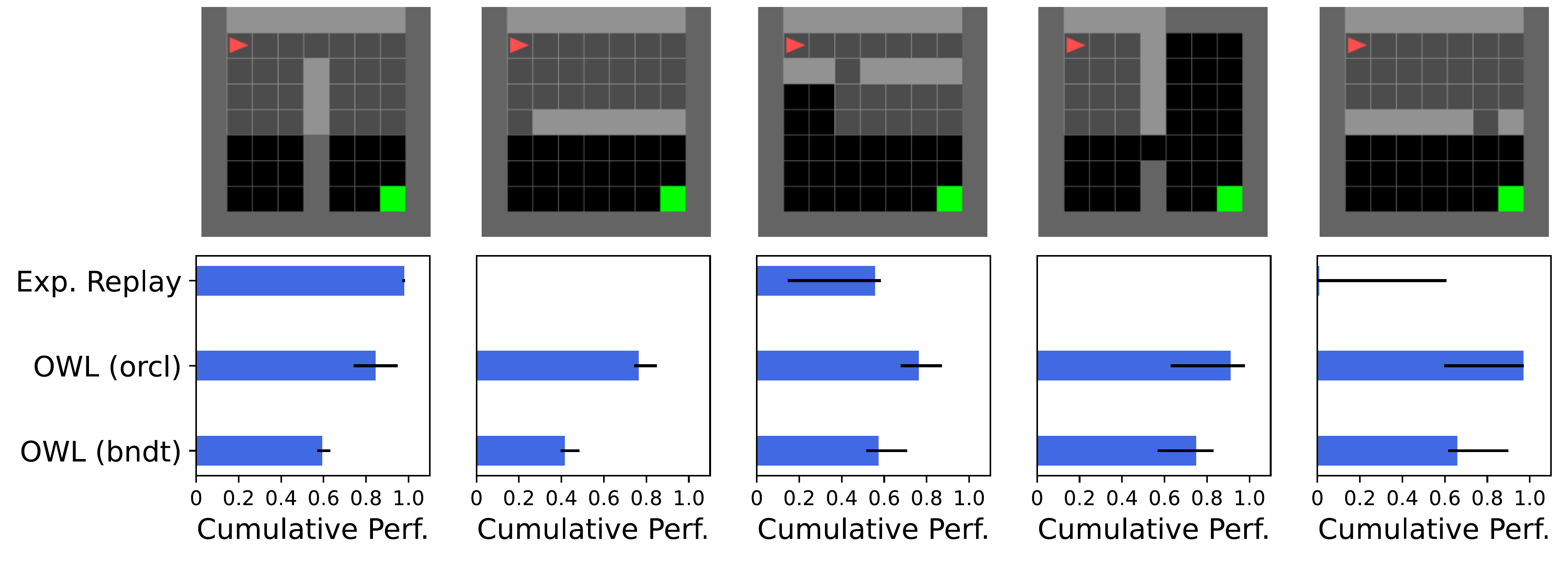} 
    \caption{Cumulative level completions for each different level for different CL strategies. Experience replay suffers from interference and cannot complete some tasks altogether.}
    \label{fig:minigrid_t5_sc}
\end{figure*}

\begin{figure*}[h]
    \hspace*{-1cm}
    \centering
    \includegraphics[width=0.95\linewidth]{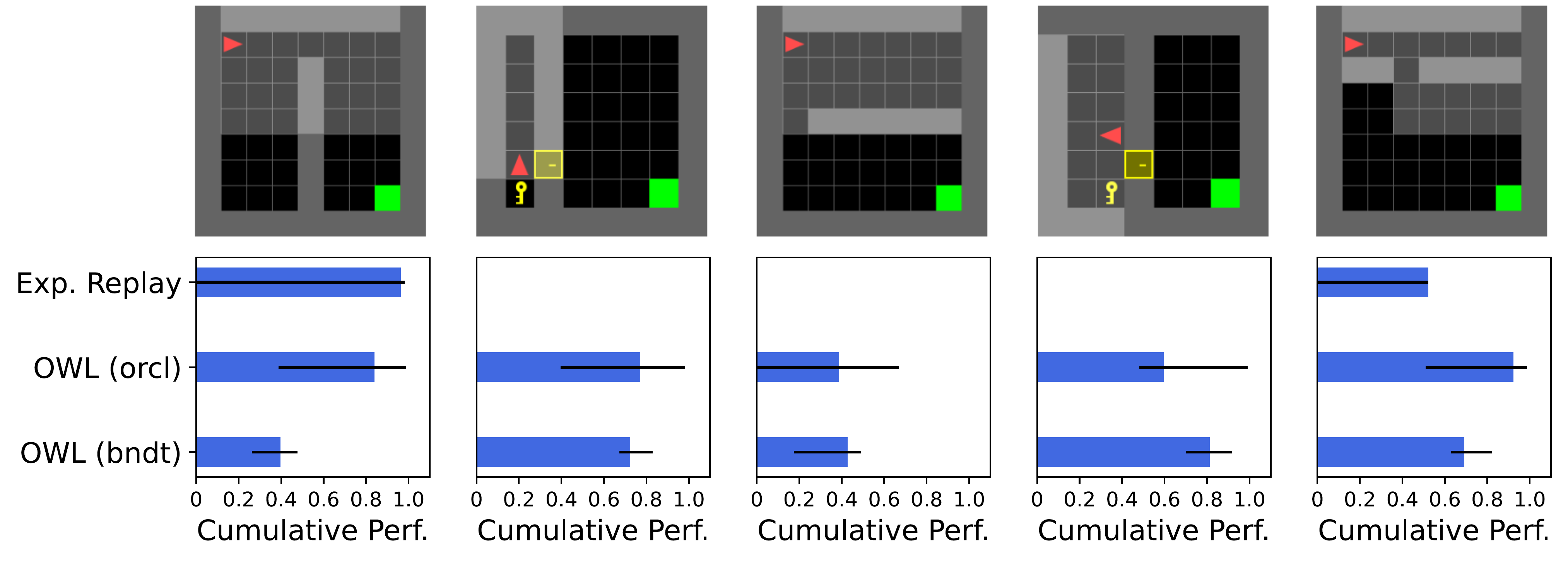} 
    \caption{Cumulative level completions for each different level for different CL strategies. Experience replay suffers from interference and cannot complete some tasks altogether.}
    \label{fig:minigrid_t5_sc_dk}
\end{figure*}

We scale to $5$ \texttt{SimpleCrossing} tasks (denoted as SC in plots) and another set of $5$ tasks where $3$ are from \texttt{SimpleCrossing} and $2$ from \texttt{DoorKey} (denoted as SC+DK). The set of tasks are repeated $3$ times each task is seen for $0.75$M environment steps. For Exp Replay we adjust the buffer size to $4$M ensure that data from all tasks are in the buffer over the course of training. Again we see that agents trained with Exp Replay suffer from interference as some tasks are never solved over the course of training, Figure~\ref{fig:minigrid_t5_sc} and Figure~\ref{fig:minigrid_t5_sc_dk}. In contrast, our OWL agents are able to solve all tasks in the face of interference, the OWL agents only see each task $20\%$ of the time but they can solve the tasks for a larger proportion of the time showing that the EWC regularization is mitigating forgetting in the feature extractor.


\section{Full Rehearsal}
\label{sec:full_rehearsal}

{\centering
\begin{minipage}{.99\linewidth}
    \centering\begin{algorithm}[H]
    \textbf{Input:} Current task $\tau \in \mathcal{T} = \{1, \ldots, M\}$, experience replay buffers per task $\left\{ \mathcal{D}[\tau], \, \forall \tau \in \mathcal{T} \right\}$. \\
    1. Sample $r \sim \textrm{U}[0, 1]$. \;\\
    \If{$r > 0.25$}{
        2. Use head $\tilde{\tau} = \tau$. \;
    }
    \Else{
        3. Use head $\tilde{\tau} \sim \textrm{U}\left[\mathcal{T} \setminus \{\tau\} \right ]$. \;
    }
    4. Sample experiences from replay buffer $\mathcal{D}[\tilde{\tau}]$. \; \\
    5. Optimize Q-function with head $\tilde{\tau}$.
     \caption{Full rehearsal: training}
    \label{alg:fr_training}
    \end{algorithm}
\end{minipage}
}

We use a multi-head network with separate replay buffers as an upper bound to OWL, named Full Rehearsal (FR). FR does not scale to many tasks, but can achieve good performance as it essentially mimics performing DQN on all tasks simultaneously while learning continuously. While training on the current task $\tau$ we need to sample a different task which is not $\tau$ every so often to ensure that these past tasks are not forgotten by the multi-headed DQN agent. Thus we devise the following simple algorithm which randomly selects a past task with probability $0.25$ to train on past tasks. This is summarized in Algorithm~\ref{alg:fr_training}. This achieves close to optimal performance, see Table~\ref{tab:minigrid_ablations_main_paper}.

\section{Bandit algorithm visualization}
\begin{figure*}
    \centering
    \includegraphics[width=0.93\linewidth]{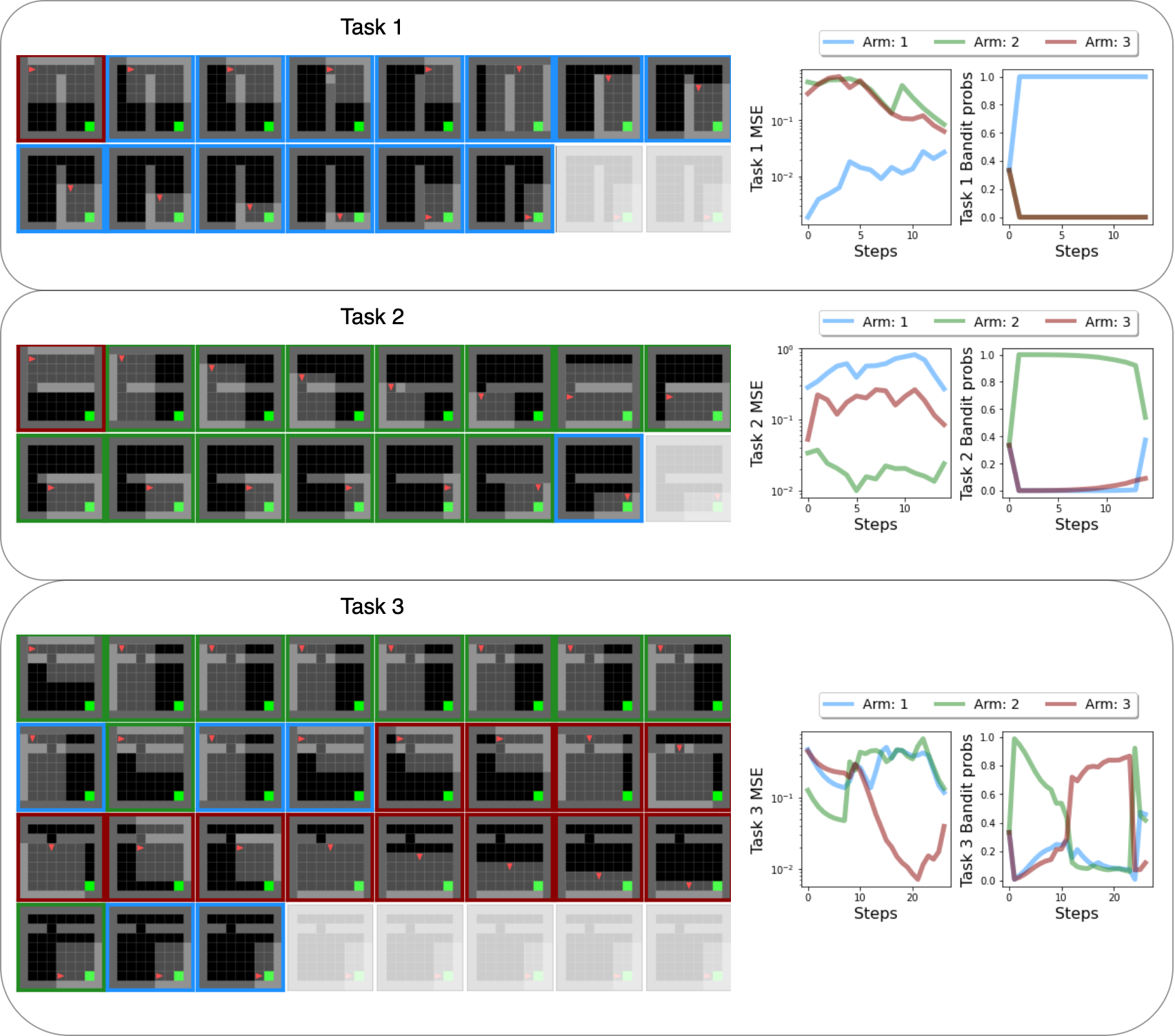} 
    
    \caption{Visualization of the OWL agent using the bandit algorithm to decide which policy to use at test time (best viewed in color). \textbf{Left}, visualization of environment and agent. The agent proceeds from left to right over a rollout. The bandit explores the different polices to begin with but then settles on the correct policy to exploit before finding the goal. \textbf{Right}, TD error (referred to as MSE in the plots) for each policy - which is fed back as the loss to the bandit algorithm and the bandit algorithm arm (policy) probabilities over the course of the rollout. For all test environments the bandit algorithm is able to find the correct policy and exploit it to find the goal. The color around the image indicates which policy / bandit arm that has been pulled. We find that close to the goal all policies give a good estimate of their Q-value and hence a similar TD errors and hence similar bandit arm probabilities. This behaviour is expected as all policies should be able to get to the goal when close by. Hence at the end of the rollout the bandit algorithm can choose a different policy head to the one that been trained for the task under evaluation, and navigate to the goal successfully.}
    \label{figure:bandit_gif}
\end{figure*}

\end{document}